\documentclass[journal]{IEEEtran}
\usepackage{ifpdf}

\usepackage{booktabs,amsfonts}
\usepackage{color}
\usepackage{array}
\usepackage{multirow}

\newcommand{\etal}{\textit{et al}.}
\newcommand{\ie}{\textit{i}.\textit{e}., }
\newcommand{\eg}{\textit{e}.\textit{g}., }

\usepackage{cite}

\ifCLASSINFOpdf
   \usepackage[pdftex]{graphicx}
\else
\fi
\usepackage{amsmath}
\usepackage{algorithmic}

\usepackage{array}
\usepackage{bbm}
\begin{document}
%
\title{Seeking Subjectivity in Visual Emotion \\ Distribution Learning}
%
%
%

\author{Jingyuan~Yang,
	~Jie~Li,
    ~Leida~Li,~\IEEEmembership{Member,~IEEE},
	Xiumei~Wang,
	Yuxuan~Ding,
	and~Xinbo~Gao,~\IEEEmembership{Senior~Member,~IEEE}
	\vspace{-10pt}
	\thanks{
		
	Manuscript received January 13, 2022; revised May 3, 2022 and July 11, 2022; accepted July 15, 2022. 
	This work was supported in part by the National Natural Science Foundation of China under Grants No. 62036007, 62176195, U21A20514, 62171340, and 61972305, in part by the Special Project on Technological Innovation and Application Development under Grant No.cstc2020jscx-dxwtB0032, and in part by Chongqing Excellent Scientist Project under Grant No. cstc2021ycjh-bgzxm0339.
	The associate editor coordinating the review of this manuscript and approving it for publication was Prof. Zhao Zhang. (\textit{Corresponding author: Xinbo Gao.})
		
	Jingyuan Yang was graduated from the School of Electronic Engineering, Xidian University, Xi'an 710071, China, and is currently with the College of Computer Science and Software Engineering, Shenzhen University, China (e-mail: jyyang@szu.edu.cn).
		
	Jie Li, Xiumei Wang and Yuxuan Ding are with the School of Electronic Engineering, Xidian University, Xi'an 710071, China (e-mail: leejie@mail.xidian.edu.cn; wangxm@xi-dian.edu.cn; yxding@stu.xidian.edu.cn).
	
	Leida Li is with the School of Artificial Intelligence, Xidian University, Xi'an 710071, China	(e-mail: ldli@xidian.edu.cn).
		
	Xinbo Gao is with the School of Electronic Engineering, Xidian University, Xi'an 710071, China (e-mail: xbgao@mail.xidian.edu.cn), and with
	the Chongqing Key Laboratory of Image Cognition, Chongqing University of Posts and Telecommunications, Chongqing 400065, China (e-mail:gaoxb@cqupt.edu.cn).
	
}
}

%
%

\markboth{IEEE TRANSACTIONS ON IMAGE PROCESSING}%
{Shell \MakeLowercase{\textit{et al.}}: Bare Demo of IEEEtran.cls for IEEE Journals}
%



\maketitle

\begin{abstract}
Visual Emotion Analysis (VEA), which aims to predict people's emotions towards different visual stimuli, has become an attractive research topic recently.
Rather than a single label classification task, it is more rational to regard VEA as a Label Distribution Learning (LDL) problem by voting from different individuals.
Existing methods often predict visual emotion distribution in a unified network, neglecting the inherent subjectivity in its crowd voting process. 
In psychology, the \textit{Object-Appraisal-Emotion} model has demonstrated that each individual's emotion is affected by his/her subjective appraisal, which is further formed by the affective memory.
Inspired by this, we propose a novel \textit{Subjectivity Appraise-and-Match Network (SAMNet)} to investigate the subjectivity in visual emotion distribution.
To depict the diversity in crowd voting process, we first propose the \textit{Subjectivity Appraising} with multiple branches, where each branch simulates the emotion evocation process of a specific individual.
Specifically, we construct the affective memory with an attention-based mechanism to preserve each individual's unique emotional experience.
A subjectivity loss is further proposed to guarantee the divergence between different individuals.
Moreover, we propose the \textit{Subjectivity Matching} with a matching loss, aiming at assigning unordered emotion labels to ordered individual predictions in a one-to-one correspondence with the Hungarian algorithm.
Extensive experiments and comparisons are conducted on public visual emotion distribution datasets, and the results demonstrate that the proposed SAMNet consistently outperforms the state-of-the-art methods. 
Ablation study verifies the effectiveness of our method and visualization proves its interpretability.

\end{abstract}


%
\IEEEpeerreviewmaketitle

\section{Introduction}
\label{sec:introduction}

Unlike any other creature on this planet, human beings are gifted with the power to express and understand emotions.
Emotion is everywhere in our daily lives, which not only decides our joy and sorrow, but also affects our behaviors and decisions. 
Therefore, aiming at finding out people's emotional reactions towards different visual stimuli, \textit{Visual Emotion Analysis (VEA)} has recently become an emerging and attractive research topic in the computer vision community.
VEA may bring potential value to related tasks (\eg memorability prediction~\cite{sidorov2019changing,lu2020understanding}, image aesthetic assessment~\cite{li2020personality,zeng2019unified}, and social relation inference~\cite{fathi2012social,zhang2015learning}), and will have great influence on wide applications (\eg opinion mining~\cite{li2019survey,sobkowicz2012opinion}, smart advertising~\cite{holbrook1984role,mitchell1986effect}, and mental disease treatment~\cite{jiang2017learning,wieser2012reduced}).

\begin{figure}
	\centering
	\includegraphics[width=0.9\linewidth]{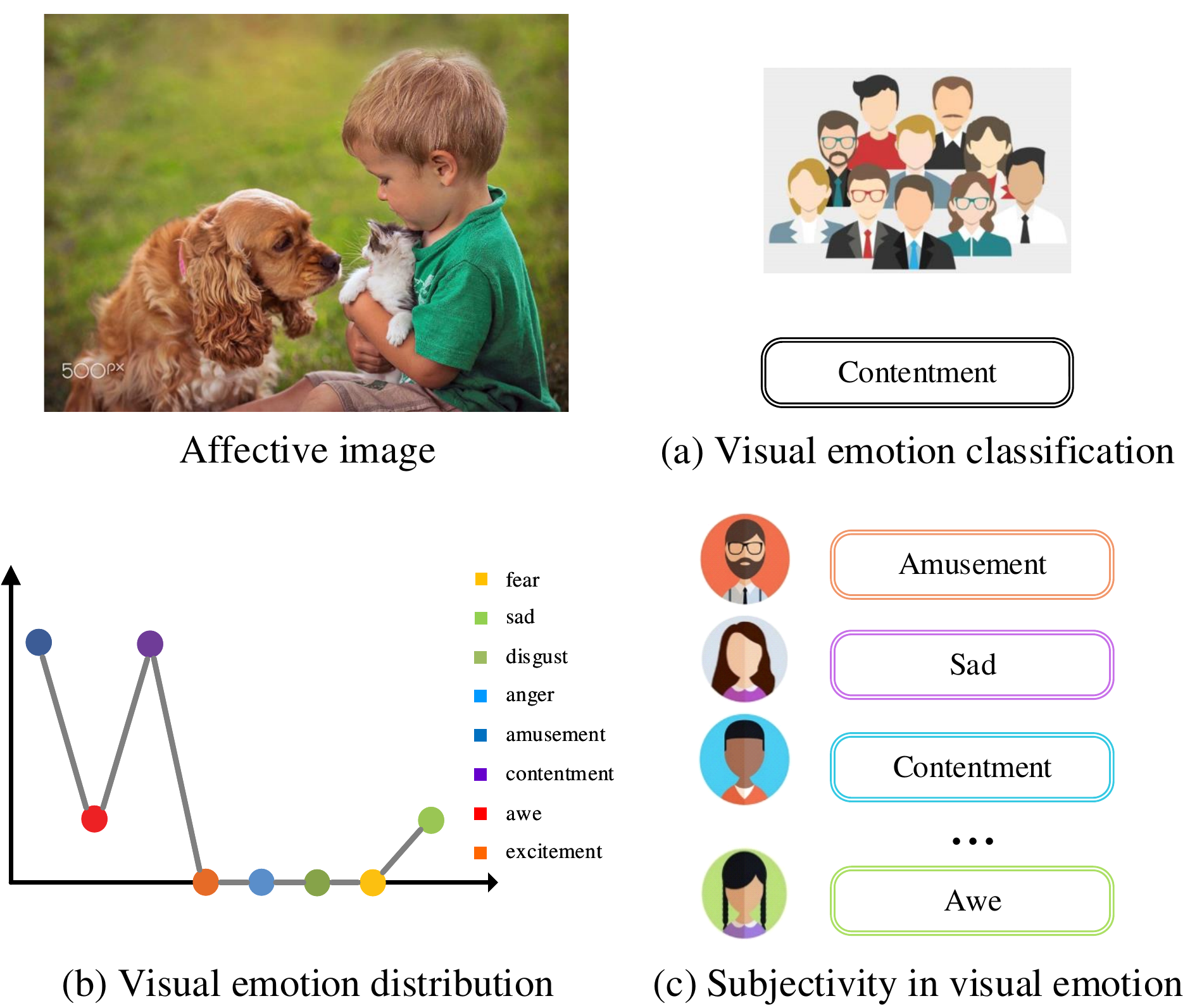}
	\caption{Exploring visual emotions from coarse to fine.
	Given an affective image, (a) assumes people have a dominant emotion, while (b) depicts crowd emotions with a label distribution.
	Since emotions are subjective, we believe that it is more rational to model the crowd voting process from an individual level, rather than a group level, as shown in (c).
	}
	\label{fig:intro_1}
\end{figure}

\textit{There are a thousand Hamlets in a thousand people's eyes.}
Similarly, there are a thousand emotions in a thousand people's eyes.
Emotions are subjective, which may vary from one individual to another.
In Fig.~\ref{fig:intro_1}(a), VEA has often been discussed as a single label classification problem,  assigning a dominant emotion label to each affective image~\cite{zhao2014exploring,yang2018weakly,yang2021solver}.
However, people's emotional experiences towards one image may be different.
Thus, it is more rational to treat VEA in a \textit{Label Distribution Learning (LDL)} paradigm, as in Fig.~\ref{fig:intro_1}(b).
Researchers have proposed several methods to address visual emotion distribution learning, which varied from the early traditional ones~\cite{geng2016label, yang2017learning, zhao2018discrete} to the recent deep learning ones~\cite{he2019image, peng2015mixed, xiong2019structured, yang2017joint, yang2017learning, yang2021circular}.
Most of the existing methods predicted the visual emotion distribution with a unified network, neglecting the subjectivity in its crowd voting process.
In reality, given an affective image, each individual is asked to vote for one emotion independently, where these votes are further summed up to form the distribution.
Therefore, as shown in Fig.~\ref{fig:intro_1}(c), we argue that it is more reasonable to take subjectivity into account, by simulating the crowd voting process in visual emotion distribution.

\begin{figure}
	\centering
	\includegraphics[width=0.8\linewidth]{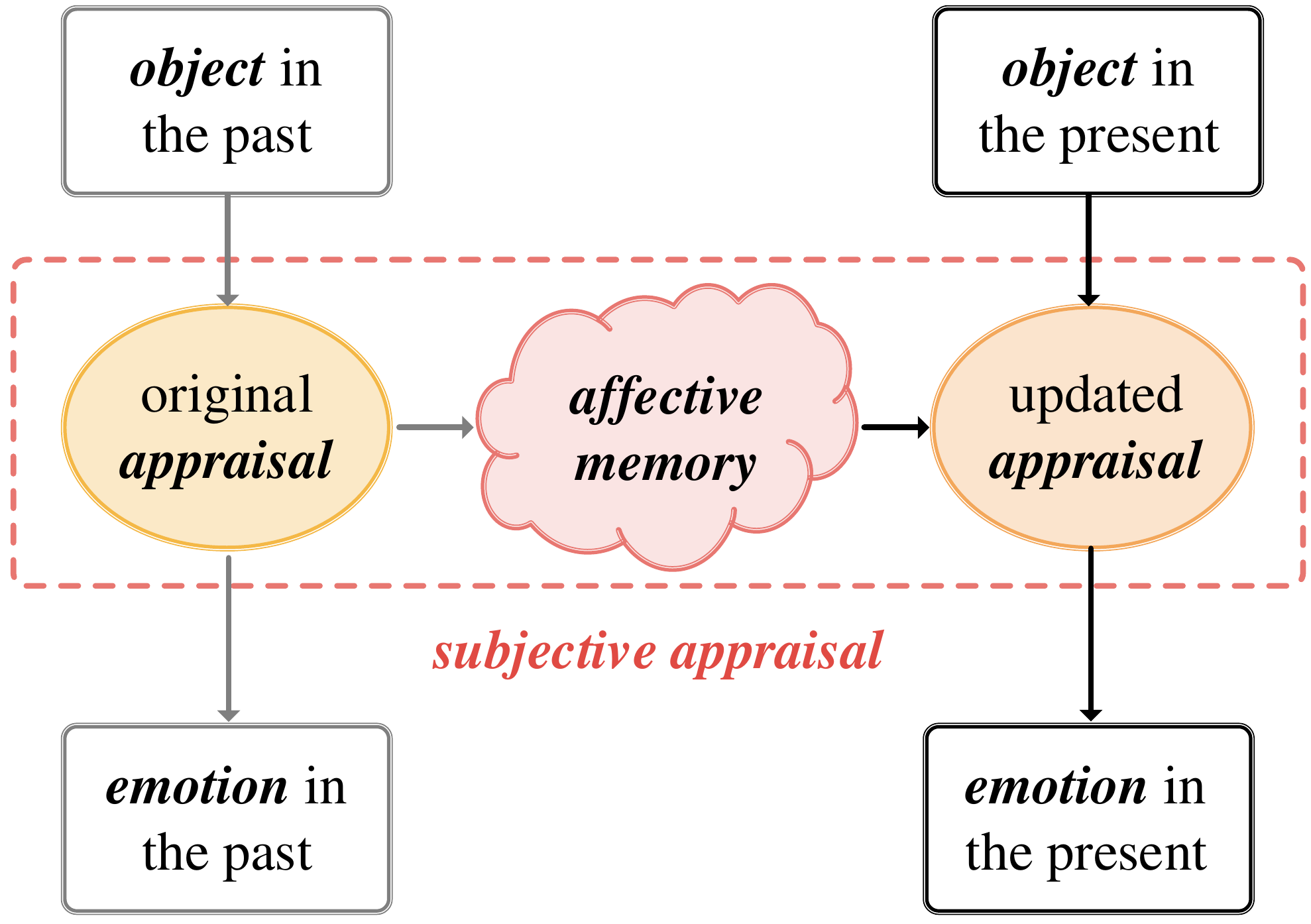}
	\caption{The \textit{Object-Appraisal-Emotion} model proposed by psychologist Arnold~\cite{mooren1993contributions}.
	Emotions are not only determined by objects, but also by appraisals, which are closely related to affective memories.
	}
	\label{fig:intro_2}
\end{figure}

In addition to computer vision, researchers in psychology~\cite{koole2009psychology, strongman1996psychology} have also devoted themselves to exploring the unknown mystery in human emotions.
As shown in Fig.~\ref{fig:intro_2}, psychologist Arnold proposed the \textit{Object-Appraisal-Emotion} model to depict the emotion evocation process, where emotions are determined by both objects and appraisals~\cite{mooren1993contributions}.
Different individuals may evoke distinct emotions towards the same object due to their subjective appraisals.
For each individual, past objects and their corresponding emotions are recorded in the affective memory, which further impacts the present emotional experience.
With more and more affective memories being stored and organized, an individual's subjective appraisal gradually formed.
Based on the above model, we believe that the diversity in visual emotion distribution is caused by subjective appraisals, which are formed by distinct affective memories of different individuals.

Therefore, we propose a novel Subjectivity Appraise-and-Match Network (SAMNet), aiming at learning visual emotion distribution in a subjective manner.
Considering the diversity in crowd emotions, we first propose the Subjectivity Appraising with multiple branches, where each represents an individual in the voting process.
Specifically, we construct an affective memory with an attention-based memory mechanism for each individual, to build connections between his/her past emotional experience and the present one.
Besides, by posing dissimilarity constraints on affective memories, we propose a subjectivity loss to guarantee the diversity between different individuals, which further preserves the subjectivity in visual emotion distribution.
Since the involved affective datasets lack individual-level annotations, there exist mismatches between individual predictions (\ie ordered) and emotion labels (\ie unordered).
In other words, given an affective image, each prediction is generated by a specific individual, while emotion labels lack individual-level annotations.
Therefore, we further propose the Subjectivity Matching with a matching loss, aiming at assigning unordered labels to ordered predictions in a one-to-one correspondence.
Specifically, we regard the problem as a bipartite matching paradigm and implement the Hungarian algorithm to find the optimal solution.

Our contributions can be summarized as follows:
\begin{itemize}
	\item We propose a novel framework, namely Subjectivity Appraise-and-Match Network (SAMNet), to seek subjectivity in visual emotion distribution learning, which outperforms the state-of-the-art methods on public visual emotion distribution datasets. 
	To the best of our knowledge, it is the initial step that predicts crowd emotions in an individual level.
	\item We construct the Subjectivity Appraising to simulate the w, where we propose affective memories to preserve individual emotional experiences and design a subjectivity loss to ensure the diversity between them.
	\item We propose the Subjectivity Matching with a matching loss, aiming for the optimal assignment between the ordered individual predictions and the unordered emotion labels by implementing the Hungarian algorithm.
\end{itemize}

The rest of the paper is organized as follows. Section~\ref{sec:related_work} overviews the existing methods on visual emotion analysis, label distribution learning, and memory networks. In Section~\ref{sec:methodology}, we introduce the proposed SAMNet by constructing Subjectivity Appraising and Subjectivity Matching. Extensive experiments, including comparisons, ablation study and visualization, are conducted on visual emotion distribution datasets given in Section~\ref{sec:experimental_results}. Finally, we conclude our work in Section~\ref{sec:conclusion}.

\section{Related work}
\label{sec:related_work}
This work addresses the problem of visual emotion analysis, which can be divided into two sub-tasks (\ie single label classification and label distribution learning).
Besides, this work is also closely related to memory networks.
This section reviews the existing methods from the above aspects.

\subsection{Visual Emotion Analysis}
\label{sec:visual_emotion_analysis}
According to different psychological models, \ie Categorical Emotion States (CES) and Dimensional Emotion Space (DES), emotions can be measured through discrete categories and continuous Valence-Arousal-Dominance space.
Based on CES model, Visual Emotion Analysis (VEA) can be grouped into single label classification and label distribution learning, where this work focuses on the latter one.
Since single label classification is the basis of label distribution learning, in this section, we thoroughly investigate both tasks in VEA.

\subsubsection{Single Label Classification}
\label{sec:single_label_classification}
Most of the earlier work in VEA focused on the design of emotional features, often inspired by the psychology and art theory~\cite{machajdik2010affective,borth2013large,zhao2014exploring,zhao2014affective}.
Machajdik~\etal~\cite{machajdik2010affective} extracted and combined specific image features, including color, texture, composition and content, from affective images to predict emotions.
Adjective Noun Pairs (ANPs) was proposed by Borth~\etal~\cite{borth2013large}, where adjectives indicated emotions while nouns corresponded to detectable objects/scenes.
To explore the relationship between emotions and artistic principles, Zhao~\etal~\cite{zhao2014exploring} extracted a series of principle-of-art-based emotional features, \ie balance, emphasis, harmony, variety, gradation, and movement.
Zhao~\etal~\cite{zhao2014affective} further proposed a multi-graph learning framework, which consisted of low-level features from elements-of-art, mid-level features from principles-of-art, as well as high-level semantic features from ANPs and facial expressions. 
Though these methods have proven effective on small-scale datasets, it is still tricky for hand-crafted features to cover all crucial features when encountering large-scale datasets.

Recently, viewing the great success of deep learning networks, researchers in VEA have adopted Convolutional Neural Network (CNN) to predict visual emotions, which also brought significant performance boost~\cite{chen2014deep,you2015robust,rao2016learning,you2017visual,yang2018visual,yang2018weakly,zhang2019exploring,yang2021stimuli,yang2021solver}.
Chen~\etal~\cite{chen2014deep} constructed DeepSentiBank based on their previous SentiBank~\cite{borth2013large}, by replacing the traditional SVM classifier with a deep CNN.
A novel progressive CNN architecture (PCNN) was proposed by You~\etal~\cite{you2015robust}, which utilized half a million images with noisy labels from the web.
Rao~\etal~\cite{rao2016learning} proposed a multi-level deep representation network (MldrNet) to predict visual emotions from multi-level, including low-level, aesthetic and semantic features.
Attention mechanism was first applied to VEA by You~\etal~\cite{you2017visual}, where emotion-relevant regions were discovered automatically.
Yang~\etal~\cite{yang2018visual} implemented off-the-shelf detection tools to find affective regions, serving as the local branch, which was further combined with the global branch to make the final prediction.
Moreover, A weakly supervised coupled network (WSCNet)~\cite{yang2018weakly} was further constructed by Yang~\etal, where emotion regions were discovered through attention mechanism in an end-to-end network.
By combining content information together with style information, Zhang~\etal~\cite{zhang2019exploring} predicted visual emotions with discriminative representations.
Recently, a stimuli-aware VEA network was proposed by Yang~\etal~\cite{yang2021stimuli}, which considered the emotion evocation process in VEA for the first time.
Further, Yang~\etal~\cite{yang2021solver} constructed a novel Scene-Object interreLated Visual Emotion Reasoning network (SOLVER), aiming at inferring visual emotions from the interactions between objects and scenes.
In reality, since emotions are ambiguous and subjective, people's emotional experiences towards one image may be different.
Besides, each emotional experience may involve multiple emotions.
Therefore, it is more reasonable to address VEA in a label distribution learning paradigm than a single label classification task.

\subsubsection{Label Distribution Learning}
\label{sec:label_distribution_learning}
Label distribution learning (LDL) was proposed by Geng~\etal~\cite{geng2016label} to assign one instance with multiple labels under different description degrees, which was regarded as a general learning framework covering single-label and multi-label classification tasks.
Traditional LDL algorithms can be grouped into three strategies, including problem transfer (PT), algorithm adaption (AA), and specialized algorithms (SA)~\cite{geng2016label}.
With the development of deep networks, Gao~\etal~\cite{gao2017deep} proposed a deep label distribution learning (DLDL) framework to prevent over-fitting by considering the label ambiguity in both feature learning and classifier learning in an end-to-end network.
Considering the inherent complexity and ambiguity lies in human emotions, LDL was introduced to VEA recently, namely visual emotion distribution learning~\cite{peng2015mixed, yang2017joint, yang2017learning, zhao2018discrete, he2019image, xiong2019structured, yang2021circular}.
For the first time, Peng~\etal~\cite{peng2015mixed} proposed to predict visual emotions in distributions rather than dominant emotions with CNNs, which is the so-called CNNR.
A novel joint classification and distribution learning network (JCDL) was constructed by Yang~\etal~\cite{yang2017joint}, which optimized softmax loss and Kullback-Leibler (KL) loss in an end-to-end network.
By adding noises to emotion distributions, Yang~\etal~\cite{yang2017learning} constructed ACPNN based on conditional probability neural networks.
With the help of graph convolutional networks, He~\etal~\cite{he2019image} proposed E-GCN to infer visual emotions from the correlations among emotions.
A method named Structured and Sparse annotations for image emotion Distribution Learning (SSDL) was proposed by Xiong~\etal~\cite{xiong2019structured}, considering the polarity and group information within emotions.
Recently, Yang~\etal~\cite{yang2021circular} constructed Emotion Circle by utilizing the intrinsic relationships between emotions from psychological models, denoted as CSR.
Most of the above methods simply predict crowd emotions collectively, neglecting the diversity in individual emotion evocation process.
In psychology, it has been demonstrated that emotions often vary from one individual to another, which is also related to their affective memories.
Therefore, we propose a Subjectivity Appraise-and-Match network (SAMNet) to seek the inherent subjectivity in visual emotion distribution.
To the best of our knowledge, it is the initial step that predicts crowd emotions in an individual level.

\subsection{Memory Networks}
\label{sec:memory_networks}
Recently, researchers have designed an external memory to augment the present network, which has been attractive due to its capability of storing and organizing the past knowledge into a structural form~\cite{weston2014memory,graves2014neural,sukhbaatar2015end,wang2018multi,pu2020learning}.
Weston~\etal~\cite{weston2014memory} first constructed a new class of learning models, namely memory networks, where long-term memory was implemented as a dynamic knowledge base.
At the same time, Grave~\etal~\cite{graves2014neural} proposed neural turing machines to extend the capabilities of neural networks, by interacting the external memory with a content-based attention mechanism.
By extending the traditional RNN network, Sukhbaatar~\etal~\cite{sukhbaatar2015end} applied a recurrent attention model over an external memory, which was trained in an end-to-end manner.
In the aforementioned methods, attention mechanisms were often implemented to construct the external memory, extracting meaningful information from the past memory.
According to the Object-Appraisal-Emotion model from psychology, affective memory is introduced to our network, which is designed to capture the subjective appraisal of different individuals.
Therefore, we implement an attention mechanism to construct our affective memory, aiming to learn present emotional experience from the past memory.

\section{Methodology}
\label{sec:methodology}
\begin{figure*}
	\centering
	\includegraphics[width=\linewidth]{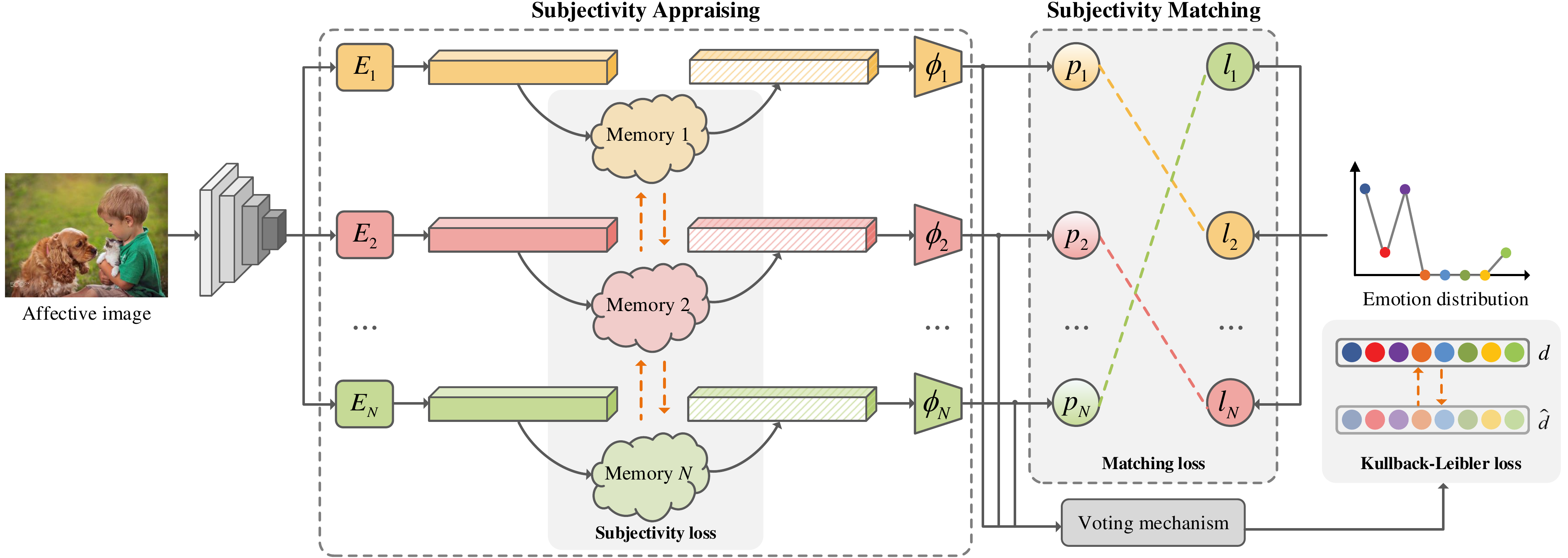}
	\caption{Framework of the proposed Subjectivity Appraise-and-Match Network (SAMNet).
		We first construct the Subjectivity Appraising with multiple branches representing different individuals, where the affective memory is designed with an attention-based mechanism to preserve each individual's emotional experience (Sec.~\ref{sec:affective_memory}).
		Besides, the subjectivity loss is proposed by posing dissimilarity constraints on different memories (Sec.~\ref{sec:subjectivity_loss}).
		In Subjectivity Matching, we construct the matching loss to find out the optimal assignment between prediction set and ground-truth set with the Hungarian algorithm (Sec.~\ref{sec:matching_loss}).
	}
	\label{fig:method}
\end{figure*}

In this section, we propose a novel Subjectivity Appraise-and-Match Network (SAMNet) to learn visual emotion distribution in a subjective manner with the guidance of the Object-Appraisal-Emotion model.
Rather than predicting emotion distribution in a unified manner, we depict different individuals with distinct affective memories to simulate the crowd voting process.
Fig.~\ref{fig:method} shows the architecture of the proposed network.
We first construct the Subjectivity Appraising, where multiple branches represent the emotion evocation process of different individuals.
For each individual, we introduce an affective memory to restore his/her unique emotional experience, which leverages an attention-based memory mechanism (Sec.~\ref{sec:affective_memory}).
A subjectivity loss is further proposed by posing dissimilarity constraints on different affective memories (Sec.~\ref{sec:subjectivity_loss}).
Moreover, we develop the Subjectivity Matching with a matching loss, aiming at assigning unordered emotion labels to ordered individual predictions in a one-to-one relation, which is modeled in a bipartite matching paradigm (Sec.~\ref{sec:matching_loss}).

\subsection{Subjectivity Appraising}
\label{sec:subjective_appraisal}

Existing methods often predict visual emotion distribution in a unified network, \ie directly output a distribution given an input image.
However, in reality, the emotion distribution is generated from the voting process of the crowd, where different people may experience different emotions.
For each affective image, several individuals are hired as voters, where each individual is asked to vote for a single emotion.
The votes from different individuals are subsequently integrated to form the final emotion distribution.
Therefore, we believe that it is more reasonable to predict emotion distribution in a subjective manner, which simulates the voting process of the crowd in real life.

In Arnold's Object-Appraisal-Emotion model, emotions are evoked by specific objects with subjective appraisals, where objects here are opposite to subjects.
Therefore, in our proposed method, we first implement a Convolutional Neural Network (CNN), \ie ResNet-50~\cite{he2016deep}, to extract the object feature ${\mathbf {f}_{obj}}$ from an affective image, where ${\mathbf {f}_{obj}} \in{{\mathbb{R}}^{d_1}}$ with $d_1=2048$.
Since a CNN backbone generally processes only the individual-agnostic, semantic-level information, the object feature extractor is shared across all individuals.
Subsequently, we construct the Subjectivity Appraising with multiple branches, where each branch depicts the subjective appraisal of a specific individual involved in the voting process.
Similar to the Object-Appraisal-Emotion model, in SAMNet, while the object feature extractor is individual-agnostic (\ie the ``object'' stage), the Subjectivity Appraising is individual-specific (\ie the ``appraisal'' stage).
It is worth mentioning that the subjective appraisal here describes each individual's unique emotional attitude towards different images.
To be specific, our network consists of $N$ branches, where each branch represents the subjective appraisal of a specific individual $n\in \left\{ 1,2,...,N \right\}$.
In order to guarantee the feature diversity among different individuals, ${\mathbf {f}_{obj}}$ is first embedded with $N$ sets of unshared parameters $E=\left\{ {{E}_{1}},{{E}_{2}},...,{{E}_{N}} \right\}$, where each ${{E}_{n}}$ is designed with an FC layer and a ReLU layer, resulting in the individual feature:
\begin{align}
\label{eq:embedding}
{\mathbf{f}_{n}}={{E}_{n}}\left( {\mathbf{f}_{obj}} \right),
\end{align}
where ${\mathbf {f}_{n}} \in{{\mathbb{R}}^{d_2}}$ with $d_2=1024$ and $N$ represents the number of individuals involved in the voting process.
 
\subsubsection{Affective Memory}
\label{sec:affective_memory}
\begin{figure}
	\centering
	\includegraphics[width=0.9\linewidth]{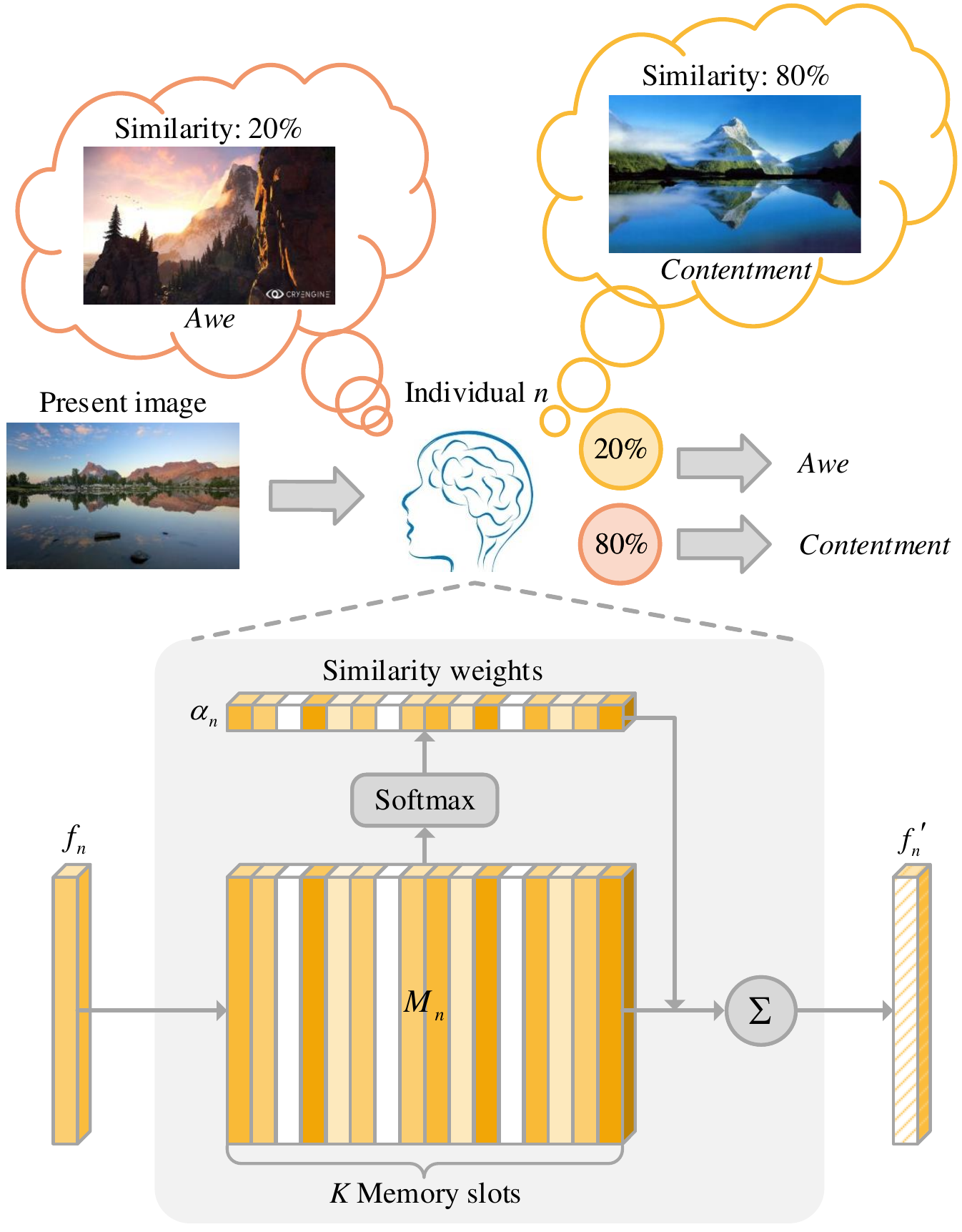}
	\caption{Affective memory module.
		By leveraging the attention-based mechanism, we construct an affective memory to store and organize the past emotional experience, which may impact the emotional experience at present.
	}
	\label{fig:mem}
\end{figure}

It has been demonstrated in Fig.~\ref{fig:intro_2} that subjective appraisals are formed by different affective memories.
Thus, we design an affective memory module to store and organize each individual's emotional experience.
Take Fig.~\ref{fig:mem} as an example, for individual $n$, images in the past (\ie mountains and lakes) and their corresponding emotions (\ie awe and contentment) are recorded in his/her affective memory, which impacts the emotional experience in the present.
When facing an image, people are likely to think of a similar image from the past emotional experience and are prone to evoke a similar emotion. 
To be specific, in Fig.~\ref{fig:mem}, the present image is $20\%$ similar to \textit{mountains} and $80\%$ similar to \textit{lakes}, resulting in the evoked emotion $20\%$ similar to \textit{awe} and $80\%$ similar to \textit{contentment}.
Motivated by the above facts, we construct the affective memory with an attention mechanism, aiming at learning the present emotion with the help of the past experience.

For each individual, the affective memory is designed as a matrix ${\mathbf {M}_{n}} \in{{\mathbb{R}}^{d_2\times K }}$, where $d_2$ is the dimension of each memory slot and $K$ denotes the number of memory slots.
In our experiment, $K$ is set to 1000, which is further ablated in Sec.~\ref{sec:hyper_parameter_analysis}.
We split an affective memory to multiple memory slots, \ie ${\mathbf {M}_{n}}=\left\{ {\mathbf{m}_{n}^{1}},{\mathbf{m}_{n}^{2}},...,{\mathbf{m}_{n}^{K}} \right\}$, where each records a different emotional experience, hoping to build connections between one specific emotion and multiple related experience.
For instance, when we think of a sunny picnic trip, we may feel contentment.
Similarly, when our lovely pets come into view, we are contentment either.
Firstly, we calculate the similarity between the individual feature ${\mathbf {f}_{n}}$ and multiple memory slots ${\mathbf {m}_{n}^{k}}$, which yields the attention weights:
\begin{align}
\label{eq:mem_slots}
{{a}_{n}^{k}}=Softmax ({\mathbf{f}_{n}}\cdot {\mathbf{m}_{n}^{k}}),
\end{align}
where $k\in \left\{ 1,2,...,K \right\}$ and ${{a}_{n}^{k}}$ represents the similarity weights on memory slots ${\mathbf{m}_{n}^{k}}$.
We apply Softmax function to normalize each attention weight to $\left[0, 1\right]$.
The more similar the individual feature ${\mathbf {f}_{n}}$ is to a memory slot ${\mathbf{m}_{n}^{k}}$, the larger attention weight ${{a}_{n}^{k}}$ it will gain.
Aiming at selectively choosing more relevant affective memories, we then apply the calculated attention weights ${\mathbf {a}_{n}}=\left\{ {{a}_{n}^{1}},{{a}_{n}^{2}},...,{{a}_{n}^{K}} \right\}$ as guidance to fuse different memory slots:
\begin{align}
\label{eq:f_n}
{\mathbf{f}_{n}'}= {\mathbf {a}_{n}}\cdot{\mathbf {M}_{n}} =\sum\limits_{k=1}^{K}({{a}_{n}^{k}}{\mathbf{m}_{n}^{k}}),
\end{align}
where ${\mathbf {f}_{n}'} \in{{\mathbb{R}}^{d_2}}$ with $d_2=1024$ and $n\in \left\{ 1,2,...,N \right\}$.
By attending weights to all memory slots, we obtain the memory-enhanced individual feature ${\mathbf{f}_{n}'}$, where individual feature is reconstructed by affective memory module.
Therefore, as is depicted in Fig.~\ref{fig:intro_2}, our memory-enhanced individual feature ${\mathbf{f}_{n}'}$ is not only determined by visual objects, but also further enhanced by the affective memory of individual $n$.

The memory-enhanced individual feature ${\mathbf{f}_{n}'}$ is subsequently sent into the emotion classifier ${{\phi }_{n}}$ and a Softmax function successively, yielding the emotion prediction:
\begin{align}
\label{eq:softmax}
{{p_n}}\left( c\left| \mathbf f_{n}',\mathbf W_n \right. \right)=\frac{\exp \left( {\mathbf{w}_{n}^c}\mathbf f_{n}' \right)}{\sum_{c=1}^{C}{\exp \left( {\mathbf{w}_{n}^c} \mathbf f_{n}' \right)}},
\end{align}
where $C$ denotes the number of emotion categories and $\mathbf W_n\in {{\mathbb{R}}^{d_2\times C}}$ is a learnable weight matrix in each emotion classifier ${{\phi }_{n}}$.
Specifically, parameters are not shared between each emotion classifier.

\subsubsection{Subjectivity Loss}
\label{sec:subjectivity_loss}
With a multi-branch design, the deep network is prone to learn similar information from each branch.
However, for an affective image, the emotion distribution is generated by diverse emotion votes from different individuals, which suggests there lies subjectivity in crowd voting process.
From Fig.~\ref{fig:intro_2}, we learn that the subjectivity is mainly attributed to different emotional experiences, \ie affective memories.
Therefore, we propose a subjectivity loss to pose constraints on different affective memories, which further ensures the emotional diversity of the crowd.
Our first thought is to calculate the similarity between any two affective memories, with the lower, the better.
In reality, however, emotions are not entirely different from one to another, \ie there may exist some commonness among different people.
Thus, we reduce the constraints by calculating the similarity between each affective memory ${\mathbf {M}_{n}}$ and the averaged affective memory $\overline{{\mathbf{M}_{n}}}$, aiming at seeking some diversity on different emotional experiences:

\begin{align}
\label{eq:m_n}
\overline{\mathbf{M}_{n}}\!=\!\frac{1}{N}\!\sum\limits_{n=1}^{N}{{\mathbf{M}_{n}}}\!=\!\left[
\!\begin{matrix}
{\frac{1}{N}\!\sum\limits_{n=1}^{N}\!{m}^{1,1}_{n}} &\cdots & {\frac{1}{N}\!\sum\limits_{n=1}^{N}\!{m}^{1,K}_{n}}      \\
\vdots & \ddots & \vdots \\
{\frac{1}{N}\!\sum\limits_{n=1}^{N}\!{m}^{d_2,1}_{n}} & \cdots & {\frac{1}{N}\!\sum\limits_{n=1}^{N}\!{m}^{d_2,K}_{n}}      \\
\end{matrix}\!
\right],
\end{align}
where $K$ and $d_2$ denote each affective memory's size, and $N$ represents the number of the involved individuals.
We first calculate the square difference between each affective memory ${\mathbf {M}_{n}}$ and the averaged one $\overline{{\mathbf{M}_{n}}}$, hoping to preserve subjectivity in crowd voting process.
After that, we use the averaged $\overline{{\mathbf{M}_{n}}}$ to normalize the square difference.
Since subjectivity loss is designed to encourage diversity, we negate the formula mentioned above and add a constant to guarantee its numerical range.
Finally, losses from all the different affective memories are summed up as a whole, namely subjectivity loss:

\begin{align}
\label{eq:sub_loss}
{{\mathcal{L}}_{sub}}\!=\!\frac{1}{N}\sum\limits_{n=1}^{N}\!{\!\left(\!\! 1\!-\!\frac{\sum\limits_{k=1}^{K}\!{\sum\limits_{l=1}^{{{d}_{2}}}{{{\!\left(\!\varphi ( m_{n}^{l,k})\!-\!\frac{1}{N}\!\sum\limits_{n=1}^{N}{\!\varphi ( m_{n}^{l,k})} \!\right)\!}^{2}}}}}{\sum\limits_{k=1}^{K}\!{\sum\limits_{l=1}^{{{d}_{2}}}{{{\left( \frac{1}{N}\!\sum\limits_{n=1}^{N}\varphi ( m_{n}^{l,k}) \right)\!}^{2}}}}} \right)},
\end{align}

\begin{align}
\label{eq:normalise}
\varphi  (\cdot)=\frac{(\cdot)-\min (\cdot)}{\max (\cdot)-\min (\cdot)}.
\end{align}

In Eq.~\ref{eq:normalise}, particularly, the affective memory ${\mathbf {M}_{n}}$ is normalized to $[0,1]$ to ensure a positive value on subjectivity loss.
With dissimilarity constraints, the subjectivity loss can ensure the diversity between different affective memories, which further preserves the subjectivity in crowd emotion voting process.

\subsection{Subjectivity Matching}
\label{sec:subjective_matching}
In previous sections, we construct a multi-branch network with affective memories to simulate the crowd voting process, which yields multiple emotion votes from different individuals.
Unfortunately, however, our affective datasets are not labeled on an individual level, \ie each emotion label lacks the specific information of the voter.
Since each emotion prediction corresponds to a specific individual (\ie ordered) while emotion labels are placed out of order (\ie unordered), we propose the Subjectivity Matching to match the unordered labels to each individual prediction in an optimal manner.

\subsubsection{Matching Loss}
\label{sec:matching_loss}
There is a kind of problem called bipartite graph matching in graph theory.
Given an undirected graph $\mathcal G=(\mathcal V,\mathcal E)$, if vertex $\mathcal V$ can be divided into two disjoint subsets $(A,B)$, and the two vertices $i$ and $j$ associated with each edge $(i,j)$ in the graph belong to these two different vertex sets $i \in A, j \in B$, then graph $\mathcal G$ is called a bipartite graph.
In a subgraph $\mathcal M$ of a bipartite graph $\mathcal G$, any two edges in the edge set ${\mathcal E}$ of $\mathcal M$ are not attached to the same vertex, then $\mathcal M$ is said to be a bipartite match.
Aiming at building a one-to-one correspondence between the disjoint prediction set and label set, our subjectivity matching problem can be modeled in a bipartite matching paradigm.
In order to compute the assignment-based set distances, various algorithms are proposed, including the widely-used Chamfer matching algorithm~\cite{barrow1977parametric} and the Hungarian algorithm~\cite{kuhn1955hungarian}, where both compare each individual prediction and its labeled counterpart and vice-versa.
We choose the Hungarian algorithm, a combinatorial optimization with high time efficiency, to solve the emotion label assignment problem.
The optimal assignment is efficiently computed with the Hungarian algorithm, where $\widehat{\sigma }$ represents the optimal matching between the prediction set and the label set:
\begin{align}
\label{eq:hun}
\widehat{\sigma }=\underset{\sigma \in {{\mathcal M}_{N}}}{\mathop{\arg \min }}\,\sum\limits_{n=1}^{N}{{{\mathcal{Q}}_{cost}}\left( {{l}_{n}},{{p}_{\sigma(n)}} \right)},
\end{align}
where $N$ denotes the number of individuals in the voting process.
In Eq.~\ref{eq:hun}, for individual $n$, $l_n$ denotes the emotion label while $p_{\sigma(n)}$ represents the matched emotion prediction.
${\mathcal{Q}}_{cost}$ is a pair-wise matching cost between ground-truth label $l_n$ and the predicted $p_{\sigma(n)}$ with index $\sigma(n)$, which is further described in Eq.~\ref{eq:cost_fun}.
Particularly, ${\mathcal M}_{N}$ denotes a set containing all the possible matching in two $N$ vertex subgraphs with a one-to-one correspondence, where $\sigma$ represents one matching within it.
The matching cost ${{\mathcal{L}}_{match}}$ is defined in a negative log-likelihood function:
\begin{align}
\label{eq:cost_fun}
{{\mathcal{Q}}_{cost}}=-{\mathbbm{1}_{\{{c=l_n}\}}}{{\widehat{p}}_{\sigma \left( n \right)}}\left( {{c}} \right),
\end{align}
where $l_n$ denotes the emotion label and $c$ is the position in the predicted vector ${\widehat{p}}_{\sigma \left( n \right)}$.
Only when $c=l_n$, \ie individual $n$ predicts emotion with the correct position in consistency with the ground-truth $l_n$, will the indicator have a non-zero value.
After finding out the optimal matching $\widehat{\sigma }$, we propose a matching loss to constrain the network from the individual level further.
To be specific, we calculate the Cross-Entropy loss between each matching pair and sum them up to form the matching loss:
\begin{align}
\label{eq:mat_loss}
{{\mathcal{L}}_{mat}}=-\frac{1}{I}\frac{1}{N}\sum\limits_{i=1}^{I}\sum\limits_{c=1}^{C}{\sum\limits_{n=1}^{N}{y_n\left( {{i,c}} \right)\log {{p}_{{\hat{\sigma }(n)}}}\left( {{i,c}} \right)}},
\end{align}
where $I$ denotes the number of affective, $C$ is the number of emotion categories, and $N$ represents the number of the involved individuals in our dataset.
Specifically, $y_n$ is the one-hot vector of the ground-truth $l_n$, ${{\hat{\sigma }(n)}}$ is the optimal matching for individual $n$.
With the proposed matching loss, our network is able to learn visual emotion distribution from the more fine-grained individual level, which also helps to investigate the subjectivity that lies in crowd emotion voting process.
For label distribution learning, it is most commonly to implement the Kullback-Leibler loss~\cite{kullback1951information} to learn the information loss caused by the inconsistency between the predicted distribution and the labeled one.
Thus, we sum up all the emotion votes from different individuals as the predicted emotion distribution, corresponding to the same mechanism in the voting process:
\begin{align}
\label{eq:voting}
\widehat{d}\left( i,c \right)=\frac{1}{N}\sum\limits_{n=1}^{N}{{{p}_{\widehat{\sigma }\left( n \right)}}\left( i,c \right)},
\end{align}
\begin{align}
\label{eq:kl_loss}
{{\mathcal{L}}_{KL}}=-\frac{1}{I}\sum\limits_{i=1}^{I}\sum\limits_{c=1}^{C}{d\left(i, c \right)\ln \widehat{d}\left(i, c \right)},
\end{align}
where KL loss is further implemented on the predicted emotion distribution $\widehat{d}\left( i,c \right)$ and the ground-truth one ${d}\left( i,c \right)$, as in Eq.~\ref{eq:kl_loss}.
Together with the proposed two losses, \ie subjectivity loss ${\mathcal{L}}_{sub}$ and matching loss ${\mathcal{L}}_{mat}$, we formulate our loss function $\mathcal{L}$ as
\begin{align}
\label{eq:loss}
\mathcal{L} = {\mathcal{L}}_{sub} + {\mathcal{L}}_{mat} + {\mathcal{L}}_{KL}.
\end{align}

With the joint loss function, SAMNet is optimized in an end-to-end network through back propagation, which predicts emotion distribution efficiently and precisely.
Moreover, with the help of subjectivity loss and matching loss, we successfully investigate the subjectivity in visual emotion distribution learning.

\begin{table*}
	\centering
	\scriptsize
	\caption{Comparison with the state-of-the-art methods on Flickr\_LDL dataset.}
	\label{tab:SOTA_Flickr}
	\renewcommand\arraystretch{1.10}
	\setlength{\tabcolsep}{1.25mm}{
		\begin{tabular}{cccccccccccccccc}
			\toprule
			\toprule
			&\multicolumn{2}{c}{PT\cite{geng2016label}}&\multicolumn{2}{c}{AA\cite{geng2016label}}&\multicolumn{3}{c}{SA\cite{geng2016label,geng2013facial}}&\multicolumn{8}{c}{CNN-based\cite{peng2015mixed,gao2017deep,yang2017learning,yang2017joint,xiong2019structured,he2019image}} \\
			\cmidrule(lr){2-3}
			\cmidrule(lr){4-5}
			\cmidrule(lr){6-8}
			\cmidrule(lr){9-16}
			Measures & PT-Bayes & PT-SVM & AA-kNN & AA-BP & SA-IIS & SA-BFGS & SA-CPNN & CNNR & DLDL & ACPNN & JCDL & SSDL & E-GCN & CSR & Ours\\
			\midrule
			Chebyshev $\downarrow$ & 0.44(14) & 0.55(15) & 0.28(9) & 0.36(12) & 0.31(11) & 0.37(13) & 0.30(10) & 0.25(6) & 0.25(6) & 0.25(6) & 0.24(5) & 0.23(3) & 0.23(3) & \textbf{0.21(1)}& \textbf{0.21(1)}\\
			Clark $\downarrow$ & 0.89(15) & 0.87(14) & \textbf{0.57(1)} & 0.82(9) & 0.82(9) & 0.86(13) & 0.82(9) & 0.84(12) & 0.78(6) & 0.77(2) & 0.77(2) & 0.78(6) & 0.78(6) & 0.77(2) & 0.77(2)\\
			Canberra $\downarrow$ & 0.85(15) & 0.83(14) & \textbf{0.41(1)} & 0.75(11) & 0.75(11) & 0.82(13) & 0.74(10) & 0.73(9) & 0.70(8) & 0.70(6) & 0.70(6) & 0.69(4) & 0.69(4) & 0.68(2)& 0.68(2)\\
			KL $\downarrow$ & 1.88(14) & 1.69(13) & 3.28(15) & 0.82(11) & 0.66(8) & 1.06(12) & 0.71(10) & 0.70(9) & 0.54(6) & 0.62(7) & 0.53(5) & 0.46(4) & 0.44(3) & 0.41(2)& \textbf{0.40(1)}\\
			Cosine $\uparrow$ & 0.63(14) & 0.32(15) & 0.79(8) & 0.72(10) & 0.78(9) & 0.70(12) & 0.70(12) & 0.72(10) & 0.81(6) & 0.80(7) & 0.82(5) & 0.85(4) & 0.86(3) & 0.87(2)& \textbf{0.88(1)}\\
			Intersection $\uparrow$ & 0.49(14) & 0.29(15) & 0.64(6) & 0.53(13) & 0.60(10) & 0.56(12) & 0.60(10) & 0.62(8) & 0.64(6) & 0.62(8) & 0.65(5) & 0.68(4) & 0.69(3) & 0.71(2)& \textbf{0.72(1)}\\
			\midrule
			Average Rank $\downarrow$ & 14.3(14) & 14.3(14) & 6.67(8) & 11(12) & 9.7(10) & 12.5(13) & 10.2(11) & 9(9) & 6.3(7) & 6.2(6) & 4.7(5) & 4.2(4) & 3.7(3) & 1.8(2) & \textbf{1.3(1)}\\
			Accuracy $\uparrow$ & 0.47(14) & 0.37(15) & 0.61(6) & 0.52(12) & 0.58(10) & 0.50(13) & 0.58(10) & 0.61(6) & 0.61(6) & 60.0(9) & 0.64(5) & 0.70(3) & 0.69(4) & 0.72(2)& \textbf{0.74(1)}\\
			\bottomrule
			\bottomrule
	\end{tabular}}
\end{table*}

\begin{table*}
	\centering
	\scriptsize
	\caption{Comparison with the state-of-the-art methods on Twitter\_LDL dataset.}
	\label{tab:SOTA_Twitter}
	\renewcommand\arraystretch{1.10}
	\setlength{\tabcolsep}{1.2mm}{
		\begin{tabular}{cccccccccccccccc}
			\toprule
			\toprule
			&\multicolumn{2}{c}{PT\cite{geng2016label}}&\multicolumn{2}{c}{AA\cite{geng2016label}}&\multicolumn{3}{c}{SA\cite{geng2016label,geng2013facial}}&\multicolumn{8}{c}{CNN-based\cite{peng2015mixed,gao2017deep,yang2017learning,yang2017joint,xiong2019structured,he2019image}} \\
			\cmidrule(lr){2-3}
			\cmidrule(lr){4-5}
			\cmidrule(lr){6-8}
			\cmidrule(lr){9-16}
			Measures & PT-Bayes & PT-SVM & AA-kNN & AA-BP & SA-IIS & SA-BFGS & SA-CPNN & CNNR & DLDL & ACPNN & JCDL & SSDL & E-GCN & CSR & Ours\\
			\midrule
			Chebyshev $\downarrow$ & 0.53(14) & 0.63(15) & 0.28(8) & 0.37(12) & 0.28(8) & 0.37(12) & 0.36(11) & 0.28(8) & 0.26(6) & 0.27(7) & 0.25(4) & 0.25(4) & 0.24(3) & \textbf{0.22(1)}& \textbf{0.22(1)}\\
			Clark $\downarrow$ & 0.85(8) & 0.91(15) & \textbf{0.58(1)} & 0.89(13) & 0.86(12) & 0.89(13) & 0.85(8) & 0.84(3) & 0.84(3) & 0.85(8) & 0.83(2) & 0.84(3) & 0.85(8) & 0.84(3)& 0.84(3)\\
			Canberra $\downarrow$ & 0.77(7) & 0.88(15) & \textbf{0.41(1)} & 0.84(13) & 0.79(12) & 0.84(13) & 0.78(9) & 0.76(2) & 0.77(7) & 0.78(9) & 0.76(2) & 0.76(2) & 0.78(9) & 0.76(2)& 0.76(2)\\
			KL $\downarrow$ & 1.31(13) & 1.65(14) & 3.89(15) & 1.19(11) & 0.64(8) & 1.19(11) & 0.85(10) & 0.67(8) & 0.54(6) & 0.58(7) & 0.53(5) & 0.51(4) & 0.46(3) & 0.44(2)& \textbf{0.43(1)}\\
			Cosine $\uparrow$ & 0.53(14) & 0.25(15) & 0.82(8) & 0.71(12) & 0.82(8) & 0.71(12) & 0.75(11) & 0.82(8) & 0.83(7) & 0.84(6) & 0.85(5) & 0.86(4) & 0.87(3) & \textbf{0.89(1)}& \textbf{0.89(1)}\\
			Intersection $\uparrow$ & 0.40(14) & 0.21(15) & 0.66(6) & 0.59(10) & 0.63(9) & 0.57(12) & 0.56(13) & 0.58(11) & 0.65(7) & 0.64(8) & 0.68(5) & 0.69(4) & 0.70(3) & 0.72(2)& \textbf{0.73(1)}\\
			\midrule
			Average Rank $\downarrow$ & 11.7(13) & 14.8(15) & 6.5(7) & 11.8(12) & 9.5(10) & 12.2(14) & 10.3(11) & 6.7(8) & 6(6) & 7.5(9) & 4.2(4) & 3.8(3) & 4.8(5) & 1.8(2) & \textbf{1.5(1)}\\
			Accuracy $\uparrow$ & 0.45(14) & 0.40(15) & 0.73(8) & 0.72(10) & 0.70(11) & 0.57(13) & 0.70(11) & 0.74(6) & 0.73(8) & 0.74(6) & 0.76(4) & 0.77(3) & 0.76(4) & 0.78(2)& \textbf{0.79(1)}\\
			\bottomrule
			\bottomrule
	\end{tabular}}
	\vspace{-10pt}
\end{table*}

\section{Experimental results}
\label{sec:experimental_results}
\subsection{Datasets}
\label{sec:datasets}
We evaluate our proposed method on public visual emotion distribution datasets, including Flickr\_LDL and Twitter\_LDL~\cite{yang2017learning}, where votes generate both datasets from a fixed number of individuals.
There are a total number of 11,150 images in Flickr\_LDL while 10,045 images in Twitter\_LDL, which are built on Mikel's eight emotion space (\ie amusement, awe, contentment, excitement, anger, disgust, fear, sad).
In the voting process, eleven individuals are hired to label Flickr\_LDL and eight individuals are hired to label Twitter\_LDL.
In both datasets, only the total number of votes on different emotions are given, while the specific individual-level annotations are not provided.
The votes from different individuals are then integrated to generate the ground-truth emotion distribution for each affective image.

\subsection{Implementation Details}
\label{sec:implementation_details]}

Our object feature extractor is constructed based on ResNet-50~\cite{he2016deep}, which is pre-trained on the large-scale visual recognition dataset, \ie ImageNet~\cite{deng2009imagenet}, following the same setting as previous methods~\cite{he2019image,yang2021circular}.
In each affective memory, we use Xavier initialization~\cite{glorot2010understanding} with a uniform distribution to initialize the parameters and set the number of memory slots to 1000.
The number of the multi-branches is determined by the number of voters in different emotion distribution datasets.
Following the same configuration in~\cite{yang2017learning}, Flickr\_LDL and Twitter\_LDL are randomly split into training set (80\%) and testing set (20\%). 
In both training and testing sets, we first resize each image to 480 on its shorter side, then crop it to 448$\times$448 randomly followed by a horizontal flip~\cite{he2016deep}.
By implementing the adaptive optimizer Adam~\cite{kingma2014adam}, the whole network is trained in an end-to-end manner with KL loss, the proposed subjectivity loss and matching loss.
With a weight decay of 5e-5, the learning rate of Adam starts from 1e-5 and is decayed by 0.1 every 10 epochs, where the total epoch number is set to 50.
Our framework is implemented using PyTorch~\cite{paszke2017automatic} and our experiments are performed on an NVIDIA GTX 1080Ti GPU.

\begin{table*}
	\centering
	\scriptsize
	\caption{Ablation study on network structure and loss functions on Flickr\_LDL dataset.
		\label{tab:ablation_flickr}
	}	
	\renewcommand\arraystretch{1.1}
	\setlength{\tabcolsep}{1.2mm}{
		\begin{tabular}{ccccccc|ccccccc}
			\toprule
			\toprule
			Single-B & Multi-B & Memory & ${\mathcal{L}}_{K\!L}$ & ${\mathcal{L}}_{sub}$ & ${\mathcal{L}}_{C\!E}$ & ${\mathcal{L}}_{mat}$ & Cheby. $\downarrow$ & Clark $\downarrow$ & Canbe. $\downarrow$ & KL $\downarrow$ & Cosine $\uparrow$ & Inter. $\uparrow$ & Acc. $\uparrow$ \\
			\midrule
			$\checkmark$&&&$\checkmark$&&&&0.239&0.783&0.697&0.435&0.843&0.678& 0.669\\
			&$\checkmark$&&$\checkmark$&&&&0.225&0.780&0.686&0.423&0.852&0.691&0.699\\
			&$\checkmark$&&$\checkmark$&&$\checkmark$&&0.217&0.781&0.687&0.412&0.868&0.701&0.698\\
			&$\checkmark$&&$\checkmark$&&&$\checkmark$&0.218&0.780&0.686&0.411&0.868&0.701&0.702\\
			&$\checkmark$&$\checkmark$&$\checkmark$&&&&0.211&0.781&0.687&0.406&0.873&0.713&0.713\\
			&$\checkmark$&$\checkmark$&$\checkmark$&&$\checkmark$&&0.209&0.779&0.684&0.409&0.874&0.717&0.717\\
			&$\checkmark$&$\checkmark$&$\checkmark$&&&$\checkmark$&0.209&0.780&0.688&0.404&0.876&0.718&0.731\\
			&$\checkmark$&$\checkmark$&$\checkmark$&$\checkmark$&&&0.209&0.778&0.683&\textbf{0.400}&\textbf{0.877}&0.718&0.725\\
			&$\checkmark$&$\checkmark$&$\checkmark$&$\checkmark$&$\checkmark$&&0.210&0.779&0.684&0.408&0.875&0.717&0.721\\
			&$\checkmark$&$\checkmark$&$\checkmark$&$\checkmark$&&$\checkmark$&\textbf{0.208}&\textbf{0.774}&\textbf{0.682}&0.404&\textbf{0.877}&\textbf{0.719}&\textbf{0.735}\\
			\bottomrule
			\bottomrule
	\end{tabular}}
\end{table*}

\begin{table*}
	\centering
	\scriptsize
	\caption{Ablation study on network structure and loss functions on Twitter\_LDL dataset.
		\label{tab:ablation_twitter}
	}	
	\renewcommand\arraystretch{1.1}
	\setlength{\tabcolsep}{1.2mm}{
		\begin{tabular}{ccccccc|ccccccc}
			\toprule
			\toprule
			Single-B & Multi-B & Memory & ${\mathcal{L}}_{K\!L}$ & ${\mathcal{L}}_{sub}$ & ${\mathcal{L}}_{C\!E}$ & ${\mathcal{L}}_{mat}$ & Cheby. $\downarrow$ & Clark $\downarrow$ & Canbe. $\downarrow$ & KL $\downarrow$ & Cosine $\uparrow$ & Inter. $\uparrow$ & Acc. $\uparrow$ \\
			\midrule
			$\checkmark$&&&$\checkmark$&&&&0.259&0.861&0.797&0.464&0.848&0.686& 0.744\\
			&$\checkmark$&&$\checkmark$&&&&0.241&0.854&0.785&0.456&0.864&0.691&0.769\\
			&$\checkmark$&&$\checkmark$&&$\checkmark$&&0.240&0.851&0.781&0.452&0.872&0.703&0.774\\
			&$\checkmark$&&$\checkmark$&&&$\checkmark$&0.241&0.852&0.779&0.449&0.875&0.708&0.777\\
			&$\checkmark$&$\checkmark$&$\checkmark$&&&&0.229&0.851&0.776&0.441&0.884&0.720&0.781\\
			&$\checkmark$&$\checkmark$&$\checkmark$&&$\checkmark$&&0.226&0.851&0.775&0.445&0.886&0.722&0.785\\
			&$\checkmark$&$\checkmark$&$\checkmark$&&&$\checkmark$&0.223&0.850&0.773&0.442&0.885&0.724&0.786\\
			&$\checkmark$&$\checkmark$&$\checkmark$&$\checkmark$&&&0.225&0.847&0.772&0.440&0.884&0.722&0.784\\
			&$\checkmark$&$\checkmark$&$\checkmark$&$\checkmark$&$\checkmark$&&0.222&0.846&0.769&0.440&0.885&0.725&0.786\\
			&$\checkmark$&$\checkmark$&$\checkmark$&$\checkmark$&&$\checkmark$&\textbf{0.220}&\textbf{0.843}&\textbf{0.762}&\textbf{0.434}&\textbf{0.887}&\textbf{0.729}&\textbf{0.790}\\
			\bottomrule
			\bottomrule
	\end{tabular}}
\end{table*}

\subsection{Comparison with the State-of-the-art Methods}
\label{sec:comparison_to_state_of_the_art}
To evaluate the effectiveness of the proposed SAMNet, we conduct experiments compared with the state-of-the-art methods on public visual emotion distribution datasets, including Flickr\_LDL and Twitter\_LDL, as shown in TABLE~\ref{tab:SOTA_Flickr} and TABLE~\ref{tab:SOTA_Twitter} respectively.
In general, the state-of-the-art visual emotion distribution learning methods can be grouped into four types: PT, AA, SA, and CNN-based methods.

\begin{itemize}
	\setlength{\itemsep}{0pt}
	\setlength{\parsep}{0pt}
	\setlength{\parskip}{0pt}
	\item \textit{Problem Transformation (PT)}:
	Aiming at transferring the LDL problem to a single-label learning (SLL) task, the PT methods, including PT-Bayes and PT-SVM, are designed based on Bayes classifier and SVM~\cite{geng2016label}.  
	Since the problem has been simplified from a label distribution (\ie multiple elements with different weights) to a single label (\ie one element), the complexity and ambiguity in label distribution are neglected, which may lead to performance drops.
	
	\item \textit{Algorithm Adaptation (AA)}:
	Traditional machine learning algorithms, \ie k-NN and BP neural network, can be extended to solve the LDL problem, which are denoted as AA-kNN and AA-BP~\cite{geng2016label}.
	As shown in TABLE~\ref{tab:SOTA_Flickr} and TABLE~\ref{tab:SOTA_Twitter}, AA-kNN maintains optimal results on Clark and Canberra in both datasets, which are difficult to surpass even for CNN-based methods.
	On the one hand, this may be attributed to the similarity between the optimization process of AA-kNN and the calculation formula of these two evaluation metrics.
	On the other hand, AA-kNN is capable of dealing with overlapped samples in visual emotion datasets.
	
	\item \textit{Specialized Algorithm (SA)}:
	Different from the previous methods, SA is designed to directly solve the LDL problem, which includes SA-IIS, SA-BFGS, SA-CPNN~\cite{geng2016label, geng2013facial}.
	Based on the previous Improved Iterative Scaling (IIS) algorithm, SA-IIS is developed to accelerate the optimization process with high efficiency~\cite{geng2016label, geng2013facial}.
	SA-BFGS is further proposed with an effective quasi-Newton method, namely Broyden-Fletcher-Goldfarb-Shanno, to avoid explicit calculation in LDL~\cite{geng2016label}.
	By implementing a three-layer conditional probability neural network, SA-CPNN is designed as another effective label distribution learning algorithm.
	Obviously, the specially designed methods improve the performance compared with the two methods mentioned above.
	
	\item \textit{CNN-based methods (CNN-based)}:
	Thanks to the powerful representation ability of CNN networks, CNN-based methods significantly boost the performance on visual emotion distribution learning.
	CNNR~\cite{peng2015mixed}, as a pioneer work, is proposed to predict visual emotions in a distribution rather than a dominant emotion by implementing an end-to-end network, which achieves a great performance boost.
	By leveraging KL loss, DLDL~\cite{gao2017deep} is further constructed to effectively utilize the label ambiguity in both feature learning and classifier learning.
	Recently, CSR~\cite{yang2021circular} takes the emotion relationships into consideration with the proposed PC loss to predict emotion distributions in a fine-grained manner.
	
\end{itemize}

We evaluate the performance of our method with six widely-used measurements in LDL, \ie Chebyshev distance ($\downarrow$), Clark distance ($\downarrow$), Canberra metric ($\downarrow$), Kullback-Leibler (KL) divergence ($\downarrow$), cosine coefficient ($\uparrow$), and intersection similarity ($\uparrow$), following the same setting as previous methods~\cite{geng2016label, he2019image, xiong2019structured, yang2017joint, yang2017learning, yang2021circular}.
Specifically, the first four are distance measurements while the last two measure similarities, where $\downarrow$ represents the lower, the better and $\uparrow$ vice versa.
In KL divergence, we use a small value $\varepsilon ={{10}^{-10}}$ to approximate the zero value, in case of a numerical explosion.
Since the maximum values of Clark distance and Canberra metric are decided by the total number of emotions, for standardized comparisons, we divide Canberra metric by the number of emotions and Clark distance by its square root.
Apart from the above six measurements, top-1 accuracy is further introduced to evaluate classification accuracy, which degenerated the LDL problem into an SLL task.
The above analysis proves that the proposed SAMNet consistently outperforms the state-of-the-art methods on widely-used visual emotion distribution datasets, which attributes to the effectiveness in simulating the crowd emotion voting process in reality.

\subsection{Ablation Study}
\label{sec:ablation_study}
Our SAMNet is further ablated to verify the validity of each module, each loss function, and the involved hyper-parameter.
\subsubsection{Network Architecture Analysis}
\label{sec:network_architecture_analysis}
In TABLE~\ref{tab:ablation_flickr} and TABLE~\ref{tab:ablation_twitter}, we conduct the ablation study on Flickr\_LDL and Twitter\_LDL datasets respectively, in order to verify the validation of each proposed module and each loss function.
Unlike the previous VEA work, we leverage multiple branches representing different individuals to simulate the crowd emotion voting process.
Therefore, we first conduct an ablation study concerning a single branch (\ie Single-B) and multiple branches (\ie Multi-B), which suggests that multiple branches bring a performance boost compared with a single one.
The proposed SAMNet mainly consists of two stages: Subjectivity Appraising (\ie Sec.~\ref{sec:subjective_appraisal}) and Subjectivity Matching (\ie Sec.~\ref{sec:subjective_matching}), which are further ablated in this section.
For Subjectivity Appraising, the proposed affective memory (\ie Memory) and subjectivity loss (\ie ${\mathcal{L}}_{sub}$) are ablated in both tables.
It is evident that the affective memory indeed significantly improves the performance of visual emotion distribution learning, owing to its ability to preserve the emotional experiences of different individuals.
For Subjectivity Matching, we compare the traditional Cross-Entropy loss (\ie ${\mathcal{L}}_{CE}$) with the proposed matching loss (\ie ${\mathcal{L}}_{mat}$), which indicates that the matching loss uniquely assigns emotion labels to individual predictions in and efficient and optimal manner.
These detailed network designs are depicted as ablated combinations in TABLE~\ref{tab:ablation_flickr} and TABLE~\ref{tab:ablation_twitter}.
From the above analysis, we can conclude that each detailed design in SAMNet is complementary and indispensable, jointly contributing to the final performance.

\begin{table}
	\centering
	\scriptsize
	\caption{Ablation study of memory slots ($K$) on Flickr\_LDL dataset.}
	\label{tab:ablation_Flickr_mem}
	\renewcommand\arraystretch{1.1}
	\setlength{\tabcolsep}{1.3mm}{
		\begin{tabular}{ccccccccc}
			\toprule
			\toprule
			Memory slots & 0 & 10 & 100 & 200 & 500 & 1000 & 2000 & 5000\\
			\midrule
			Chebyshev $\downarrow$ & 0.218 & 0.217 & 0.215 & 0.211 & 0.209 & \textbf{0.208} & 0.210 & 0.211 \\
			Clark $\downarrow$ & 0.780 & 0.781 & 0.780 & 0.780 & 0.778 & \textbf{0.774} & 0.780 & 0.779 \\
			Canberra $\downarrow$ & 0.686 & 0.686 & 0.685 & 0.686 & 0.684 & \textbf{0.682} & \textbf{0.682} & 0.685 \\
			KL $\downarrow$ & 0.411 & 0.413 & 0.408 & 0.405 & 0.406 & \textbf{0.404} & 0.406 & 0.409 \\
			Cosine $\uparrow$ & 0.868 & 0.867 & 0.870 & 0.874 & 0.876 & \textbf{0.877} & 0.875 & 0.873 \\
			Intersection $\uparrow$ & 0.701 & 0.704 & 0.705 & 0.710 & 0.715 & \textbf{0.719} & 0.716 & 0.715 \\
			Accuracy $\uparrow$ & 0.702 & 0.705 & 0.715 & 0.718 & 0.726 & \textbf{0.735} & 0.725 & 0.724 \\
			\bottomrule			
			\bottomrule	\end{tabular}}
	\vspace{-5pt}
\end{table}

\begin{table}
	\centering
	\scriptsize
	\caption{Ablation study of memory slots ($K$) on Twitter\_LDL dataset.}
	\label{tab:ablation_Twitter_mem}
	\renewcommand\arraystretch{1.1}
	\setlength{\tabcolsep}{1.3mm}{
		\begin{tabular}{ccccccccc}
			\toprule
			\toprule
			Memory slots & 0 & 10 & 100 & 200 & 500 & 1000 & 2000 & 5000\\
			\midrule
			Chebyshev $\downarrow$ & 0.241 & 0.239 & 0.229 & 0.227 & 0.223 & \textbf{0.220} & 0.222 & 0.228 \\
			Clark $\downarrow$ & 0.852 & 0.849 & 0.846 & 0.848 & 0.845 & \textbf{0.843} & 0.845 & 0.847 \\
			Canberra $\downarrow$ & 0.779 & 0.774 & 0.771 & 0.771 & 0.768 & \textbf{0.762} & 0.769 & 0.771 \\
			KL $\downarrow$ & 0.449 & 0.451 & 0.445 & 0.441 & 0.437 & \textbf{0.434} & 0.436 & 0.438 \\
			Cosine $\uparrow$ & 0.875 & 0.874 & 0.878 & 0.879 & 0.885 & \textbf{0.887} & 0.884 & 0.883 \\
			Intersection $\uparrow$ & 0.708 & 0.709 & 0.714 & 0.722 & 0.726 & \textbf{0.729} & 0.727 & 0.725 \\
			Accuracy $\uparrow$ & 0.777 & 0.778 & 0.780 & 0.782 & 0.784 & \textbf{0.790} & 0.785 & 0.784 \\
			\bottomrule			
			\bottomrule	\end{tabular}}
	\vspace{-5pt}
\end{table}

\subsubsection{Hyper-Parameter Analysis}
\label{sec:hyper_parameter_analysis}
We conduct experiments on Flickr\_LDL and Twitter\_LDL to validate the choice of memory slots number $K=1000$ in the affective memory (Sec.~\ref{sec:affective_memory}), as shown in TABLE~\ref{tab:ablation_Flickr_mem} and TABLE~\ref{tab:ablation_Twitter_mem}.
The number of memory slots $K$ represents the capacity of emotional experience in the human brain, where a higher number may contain richer information. 
Setting $K$ from $0$ to $5000$, we evaluate the six measurements as well as the top-1 accuracy in LDL successively.
We find that most measurements constantly gain performance boost as $K$ varies from 0 to 1000,  while meeting drops slightly after $K=1000$.
The growing performance further proves that the proposed affective memory can record and organize emotional experience, which further ensures its effectiveness in visual emotion distribution learning.
Besides, it also suggests that the affective memory can be depicted in a more precise and comprehensive manner with a more significant number of memory slots.
However, the performance meets slightly drops when $K$ is too large, which may be attributed to the overfitting of too many parameters in memory slots.
The above observations indicate that a proper value of memory slots brings optimal performance, which is consistent with the human brain's memory mechanism.

\subsection{Visualization}
\label{sec:visualization}

\begin{figure*}
	\centering
	\includegraphics[width=\linewidth]{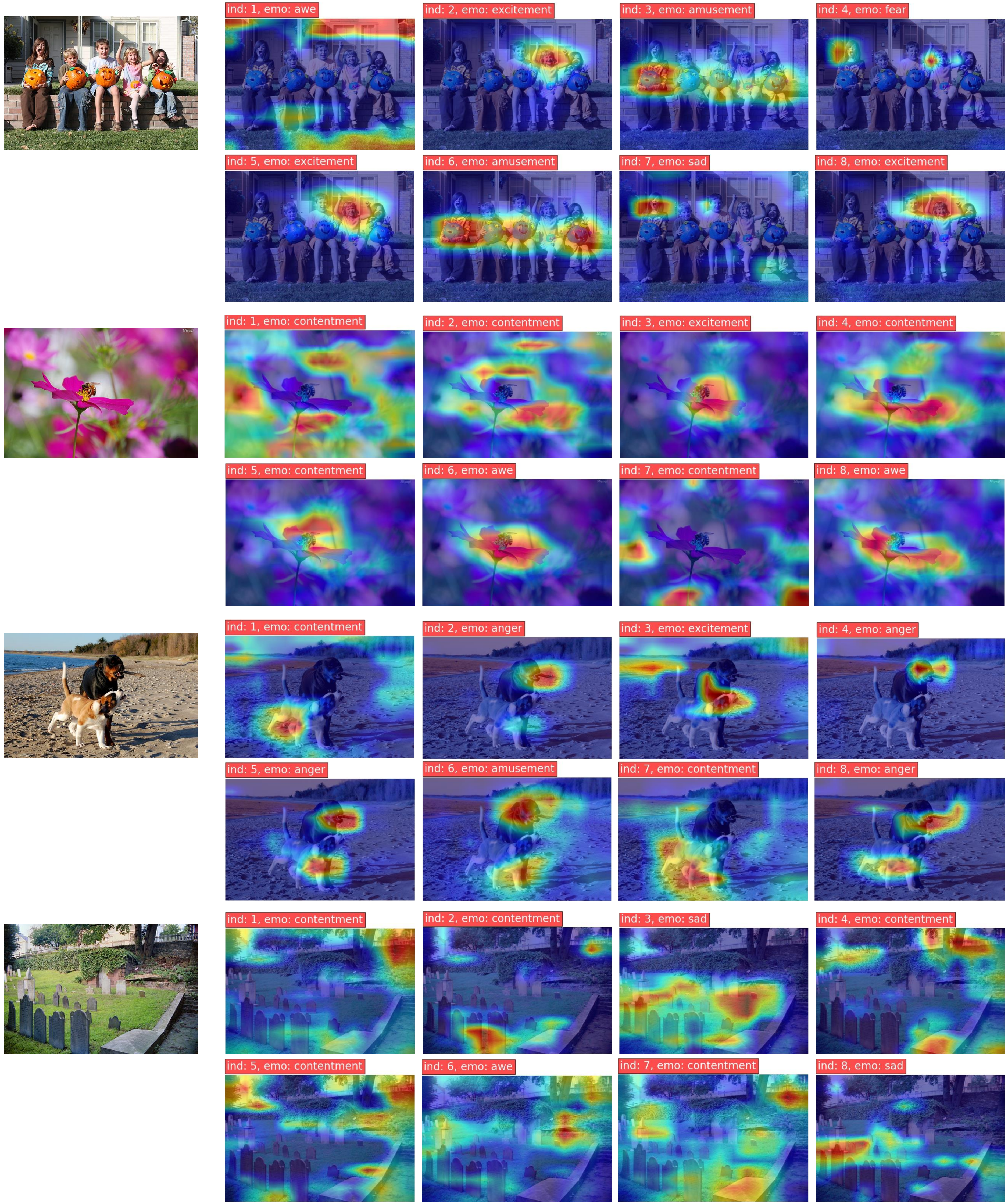}
	\caption{Visualization of affective memories from different individuals on Flickr\_LDL datasets.
		In each individual's memory map, red indicates areas that are more likely to link to his/her emotional experience and thus largely determine the evoked emotion.
		The order number of each individual and the evoked emotion are marked at the top of each memory map.
	}
	\label{fig:vis_mem_1}
\end{figure*}

\begin{figure*}
	\centering
	\includegraphics[width=\linewidth]{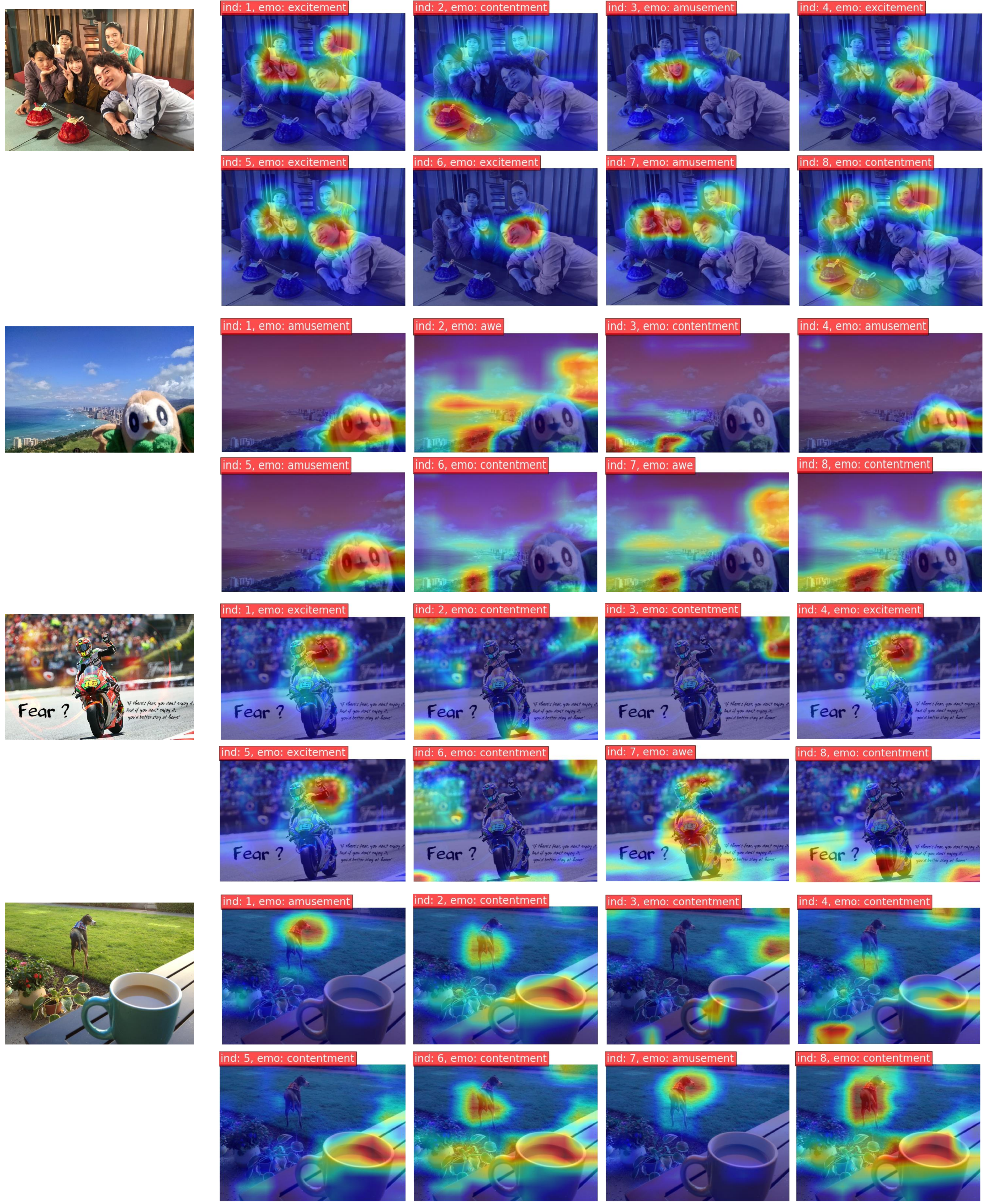}
	\caption{Visualization of affective memories from different individuals on Twitter\_LDL datasets.
		In each individual's memory map, red indicates areas that are more likely to link to his/her emotional experience and thus largely determine the evoked emotion.
		The order number of each individual and the evoked emotion are marked at the top of each memory map.
	}
	\label{fig:vis_mem_2}
\end{figure*}

\begin{figure*}
	\centering
	\includegraphics[width=\linewidth]{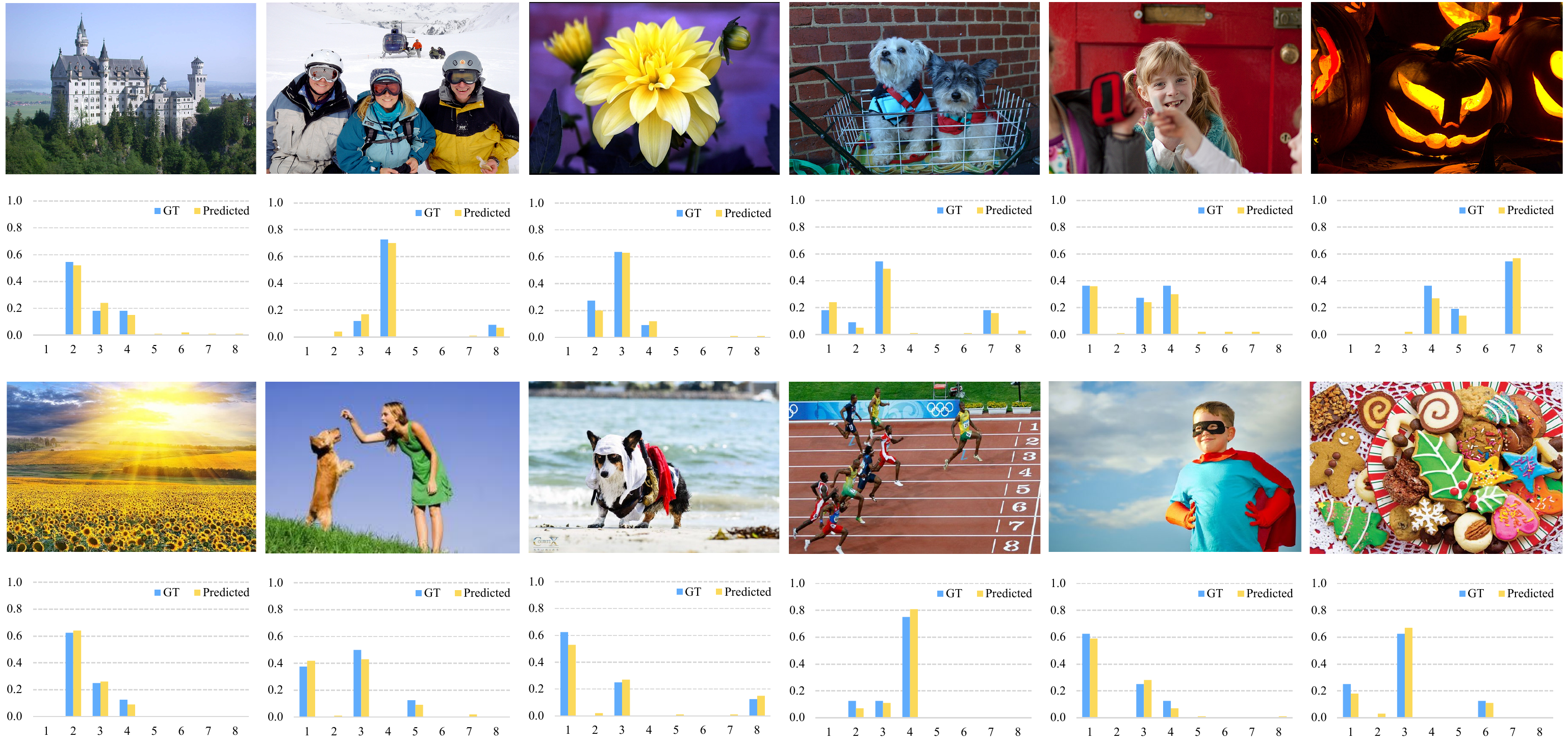}
	\caption{Visualization of the predicted emotion distributions (\ie Predicted) and the ground-truth (\ie GT) ones, where images in the first row come from Flickr\_LDL dataset and second row the Twitter\_LDL.
		Numbers on the horizontal axis correspond to the emotion categories in the involved affective datasets.
	}
	\label{fig:vis_dist}
\end{figure*}

The effectiveness and robustness of the proposed SAMNet have been quantitatively evaluated in previous sections.
Since our motivation is to simulate the crowd emotion voting process by designing affective memories, in this section, we first try to figure out how different individuals react distinctively towards the same image and how affective memories influence the evoked emotions (Sec.~\ref{sec:memory_map}).
Besides, we also visualize the predicted emotion distributions and the ground-truths, which further indicates that SAMNet is capable of predicting emotion distributions by investigating subjectivity in the crowd in a more refined manner(Sec.~\ref{sec:emotion_distribution_vis}).

\subsubsection{Memory Map}
\label{sec:memory_map}
In order to better illustrate the design of the affective memory, we visualize each memory-enhanced individual feature ${\mathbf {f}_{n}'}$ with Grad-CAM~\cite{selvaraju2017grad} mechanism, denoted as memory map.
To be specific, we first denote $\mathbf{A}^k$ as the $k$-th activation map in the output layer of the backbone CNN.
Grad-CAM uses the gradient information flowing into the last convolutional layer, which assigns importance values to each neuron for a particular decision of interest~\cite{selvaraju2017grad}.
Therefore, we calculate the gradient of the score for individual $n$, ${\mathbf {f}_{n}'}$, with respect to
$\mathbf {A}^k$, denoted as neuron importance weight $\alpha _{k}^{n}$:
\begin{align}
\label{eq:activation}
\alpha _{k}^{n}=\frac{1}{Z}\sum\limits_{i}{\sum\limits_{j}{\frac{\partial {\mathbf{f}_{n}'}}{\partial A_{ij}^{k}}}},
\end{align}
where $k$ denotes the $k$-th activation map and $n$ suggests the $n$-th individual involved in the voting process.
Moreover, $i$ and $j$ indicate the position while $Z$ is the size of the activation map.
A weighted combination of activation maps $\mathbf{A}^k$ are further computed, followed by a ReLU function successively:
\begin{align}
\label{eq:memory_map}
\mathbf {M{{M}}}_{n}=\operatorname{ReLU}(\sum\limits_{k}{\alpha _{k}^{n}{\mathbf {A}^{k}}}).
\end{align}

Thus, the memory map $\mathbf {M{{M}}}_{n}$ of individual $n$ is obtained, where the divergences of affective memories are presented with different spatial information in the heatmaps. 
Fig.~\ref{fig:vis_mem_1} shows the visualizations from Flickr\_LDL dataset while Fig.~\ref{fig:vis_mem_2} from Twitter\_LDL dataset.
The first image represents the original affective image and the rest shows the memory maps of different individuals (\ie individual1 - individual8) correspondingly.
Notably, to present the visualization results in a unified manner, we choose the first eight of the eleven individuals in Flickr\_LDL dataset, keeping up with the eight voters in the Twitter\_LDL dataset.
In Fig.~\ref{fig:vis_mem_1} and Fig.~\ref{fig:vis_mem_2}, the order number and the evoked emotion of each individual are marked at the top of each memory map.

In Fig.~\ref{fig:vis_mem_1}, take the first image as an example, where there are kids, pumpkin lanterns and a leisure courtyard.
When it comes to the laughing kids, people may memorize similar smiles with excitement.
When it comes to pumpkin lanterns, it may remind people of their Halloween experiences and thus evoke amusement.
When it comes to the leisure courtyard, one may remember his own garden and feel awe and yearning.
Similarly, there are city landscapes, the blue sky and a cartoon doll in Fig.~\ref{fig:vis_mem_2}.
The cartoon doll may remind people of something funny and amusing, while the city landscape can arouse the feeling of home, calm and contentment.
Viewing the blue sky and white clouds, one may be shocked by the beautiful scenery and hold in awe and veneration.
In other cases, ferocious fangs can bring anger, birthday cakes may remind us of contentment, and motor-racing evokes excitement, which are all related to our previous emotional experience.
From the above analysis, we can conclude that, for each individual, the affective memory differs from one to another, which is related to his/her unique emotional experience.
Besides, it is suggested that similar affective memories may evoke similar emotions, which also proves that each individual's emotion is largely determined by his/her affective memory (Sec.~\ref{sec:affective_memory}).
In Object-Appraisal-Emotion model, each individual's subjective appraisal is formed by his/her affective memories.
In this section, we visualize these affective memories with memory maps to find some diversity between different individuals.
Fig.~\ref{fig:vis_mem_1} and Fig.~\ref{fig:vis_mem_2} show the diversity in crowd emotions, which is the so-called subjectivity in visual emotion distribution.

\subsubsection{Emotion Distribution}
\label{sec:emotion_distribution_vis}
We further visualize the predicted emotion distributions and the ground-truth labels of affective images on Flickr\_LDL and Twitter\_LDL datasets.
In Fig.~\ref{fig:vis_dist}, images in the first row come from Flickr\_LDL and the second row Twitter\_LDL, where yellow represents the predicted emotion distribution (\ie Predicted) and blue the ground truth (\ie GT).
Notably, numbers (\ie 0-7) on the horizontal axis correspond to the emotion categories on Mikel's wheel (\ie amusement, awe, contentment, excitement, anger, disgust, fear, sad), where the involved affective datasets are built upon.
It is shown in Fig.~\ref{fig:vis_dist} that SAMNet is capable of accurately capturing discrepant emotional feedback from different individuals, owing to the simulation of crowd emotion voting process.
In some cases, emotional experience may differ from one individual to another to a large extent (\ie amusement and sad), SAMNet can still well depict the emotion divergence, which attributes to the design of subjectivity loss.
Besides, when crowd emotions are convergent, SAMNet can still seek common ground while reserving differences, yielding relatively good results in such cases.

\section{Conclusion}
\label{sec:conclusion}
We have proposed a Subjectivity Appraise-and-Match Network (SAMNet) to seek subjectivity in visual emotion distribution learning, by simulating the crowd emotion voting process.
We first constructed Subjectivity Appraising with the affective memory to preserve individual emotional experience, where a subjective loss was designed to ensure the diversity in emotion distributions.
Then, we proposed Subjectivity Matching with a matching loss, which assigned emotion labels to individual predictions in a bipartite matching paradigm.
Extensive experiments and comparisons have shown that the proposed SAMNet consistently outperforms the state-of-the-art methods on public visual emotion distribution datasets.
Visualization results further proved the justifiability and interpretability of the proposed affective memory, which also brought new sights to investigate subjectivity in visual emotion distribution learning.
There are also some limitations in the proposed network. 
Since SAMNet is proposed to seek subjectivity, it may neglect the similarity between different individuals. 
Therefore, how to well balance the diversity and the similarity in visual emotion distribution learning remains an important research topic. 
Besides, most of the existing visual emotion datasets lack individual-level labels and personalized information. 
With such annotations, our SAMNet can be upgraded with personalized guided affective memories as well as correct prediction-label pairs without matching strategy. 
We may consider the above limitations in our follow-up work.



\bibliographystyle{IEEEtran}
\bibliography{mybibfile}
%

%

%
%
%




\begin{IEEEbiography}[{\includegraphics[width=1in,height=1.25in,clip,keepaspectratio]{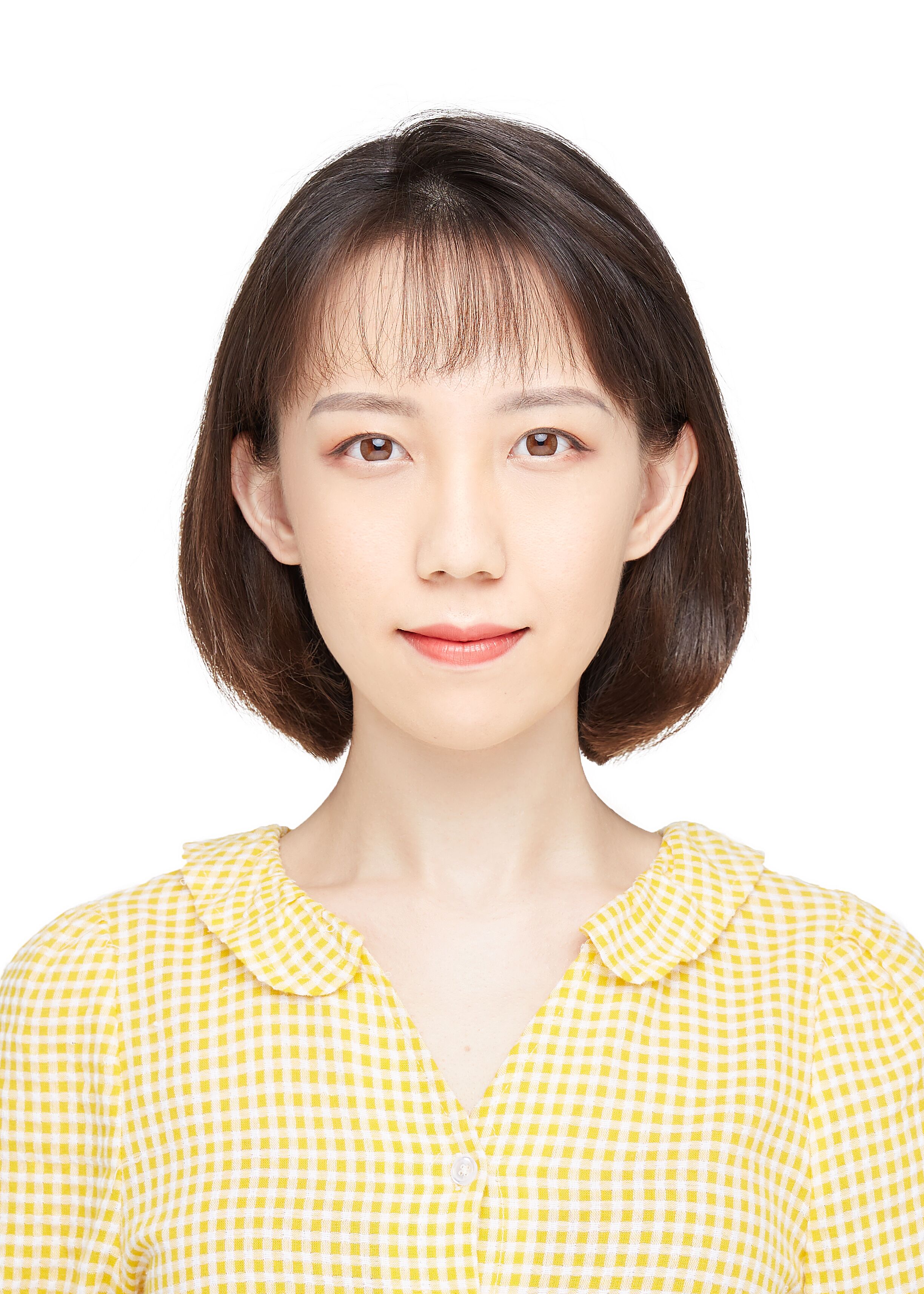}}]{Jingyuan Yang}
	received the B.Eng. and Ph.D. degrees in electronic engineering, signal and information processing from Xidian University, Xi'an, China, in 2017 and 2022. She is currently an Assistant Professor with the Visual Computing Research Center at the College of Computer Science and Software Engineering, Shenzhen University. Her current research interest is visual emotion analysis, image aesthetics and generation tasks.
\end{IEEEbiography}

\begin{IEEEbiography}[{\includegraphics[width=1in,height=1.25in,clip,keepaspectratio]{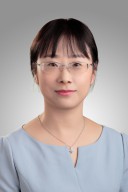}}]{Jie Li}
	received the B.Sc. degree in electronic engi-neering, the M.Sc. degree in signal and information processing, and the Ph.D. degree in circuit and sys-tems from Xidian University, Xi’an, China, in 1995, 1998, and 2004, respectively. She is currently a Professor with the School of Electronic Engineering, Xidian University. Her research interests include image processing and machine learning. In these areas, she has published around 50 technical articles in refereed journals and proceedings, including the IEEE T-IP, T-CSVT, and Information Sciences.
\end{IEEEbiography}

\begin{IEEEbiography}[{\includegraphics[width=1in,height=1.25in,clip,keepaspectratio]{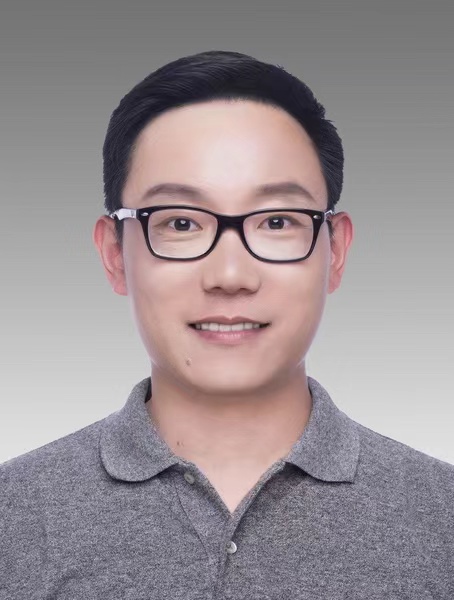}}]{Leida Li}
	(Member, IEEE) received the B.S. and Ph.D. degrees from Xidian University, Xi’an, China, in 2004 and 2009, respectively. In 2008, he was a Research Assistant with the Department of Electronic Engineering, National Kaohsiung University of Science and Technology, Kaohsiung, Taiwan. From 2014 to 2015, he was a Visiting Research Fellow with the Rapid-Rich Object Search (ROSE) Lab, School of Electrical and Electronic Engineering, Nanyang Technological University, Singapore, where he was a Senior Research Fellow, from 2016 to 2017. He is currently a Professor with the School of Artificial Intelligence, Xidian University, China. His research interests include multimedia quality assessment, affective computing, information hiding, and image forensics.
	He has served as an SPC for IJCAI 2019-2021, the Session Chair for ICMR 2019 and PCM 2015, and TPC for CVPR 2021, ICCV 2021, AAAI 2019-2021, ACM MM 2019-2020, ACM MM-Asia 2019, and ACII 2019.
	He is currently an Associate Editor of the Journal of Visual Communication and Image Representation and the EURASIP Journal on Image and Video Processing.
\end{IEEEbiography}

\begin{IEEEbiography}[{\includegraphics[width=1in,height=1.25in,clip,keepaspectratio]{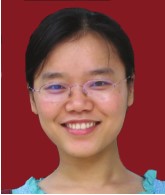}}]{Xiumei Wang}
	received the Ph.D. degree from Xidian University, Xi’an, China, in 2010. She is currently a Lecturer with the School of Electronic Engineering, Xidian University. Her current research interests include nonparametric statistical models and machine learning. She has published several scientific articles, including the IEEE Trans. Cybernetics, Pattern Recognition, and Neurocomputing in the above areas.
\end{IEEEbiography}

\begin{IEEEbiography}[{\includegraphics[width=1in,height=1.25in,clip,keepaspectratio]{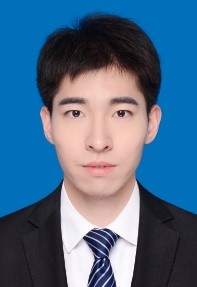}}]{Yuxuan Ding}
	was born in 1995. He received the B.Eng. degree in Intelligent Science and Technology from Xidian University, Xi’an, China, in 2018. He is currently a Ph. D. Candidate at the School of Electronic Engineering, Xidian University. His main research interest covers Machine Learning, Computer Vision, Vision-Language, and their applications.
\end{IEEEbiography}

\begin{IEEEbiography}[{\includegraphics[width=1in,height=1.25in,clip,keepaspectratio]{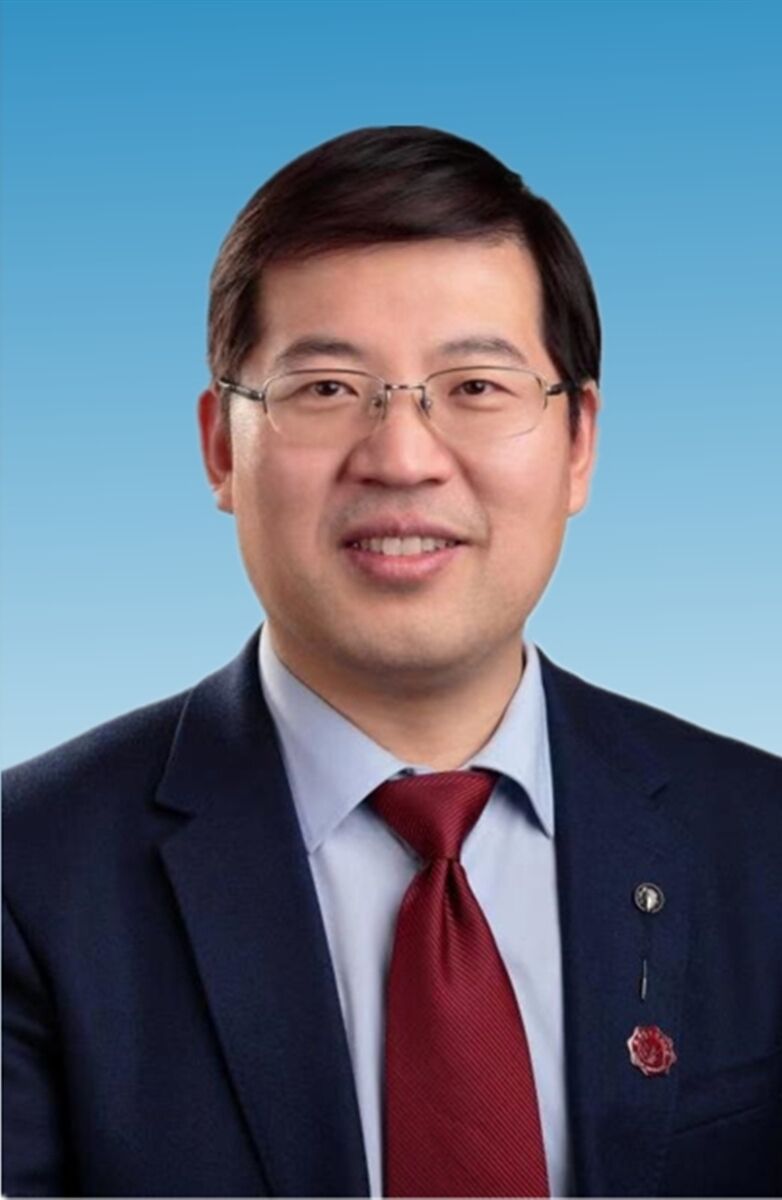}}]{Xinbo Gao}
	(Senior Member, IEEE) received the B.Eng., M.Sc. and Ph.D. degrees in electronic engineering, signal and information processing from Xidian University, Xi'an, China, in 1994, 1997, and 1999, respectively. From 1997 to 1998, he was a research fellow at the Department of Computer Science, Shizuoka University, Shizuoka, Japan. From 2000 to 2001, he was a post-doctoral research fellow at the Department of Information Engineering, the Chinese University of Hong Kong, Hong Kong. Since 2001, he has been at the School of Electronic Engineering, Xidian University. He is a Cheung Kong Professor of Ministry of Education of P. R. China, a Professor of Pattern Recognition and Intelligent System of Xidian University. Since 2020, he has been also a Professor of Computer Science and Technology of Chongqing University of Posts and Telecommunications. His current research interests include image processing, computer vision, multimedia analysis, machine learning and pattern recognition. He has published seven books and around 300 technical articles in refereed journals and proceedings. Prof. Gao is on the Editorial Boards of several journals, including Signal Processing (Elsevier) and Neurocomputing (Elsevier). He served as the General Chair/Co-Chair, Program Committee Chair/Co-Chair, or PC Member for around 30 major international conferences. He is a Fellow of the Institute of Engineering and Technology, a Fellow of the Chinese Institute of Electronics, a Fellow of the China Computer Federation, and Fellow of the Chinese Association for Artificial Intelligence.
\end{IEEEbiography}

\end{document}